\documentclass{SCIS2025}
\usepackage{layout}
\usepackage{hyperref}

\usepackage{amssymb}
\usepackage{amsmath}
\usepackage{booktabs}
\usepackage{multirow}
\usepackage{siunitx}
\usepackage{lipsum}
\usepackage{multicol}

\usepackage{url}
\usepackage{array}
\usepackage{comment}
\usepackage{tabularx}
\usepackage{listings}
\usepackage[dvipsnames]{xcolor}
\definecolor{hidden-black}{rgb}{0,0,0} 
\definecolor{hidden-green}{rgb}{0,0.5,0} 
\usepackage{fancyvrb}  

\usepackage{graphicx}    
\usepackage{tikz}
\usetikzlibrary{arrows.meta, positioning}
\usepackage[edges]{forest}
\usetikzlibrary{trees}
\usepackage[most]{tcolorbox}
\usepackage[utf8]{inputenc}

\begin{document}
\ArticleType{REVIEW}
\Year{2025}
\Month{January}
\Vol{68}
\No{1}
\DOI{}
\ArtNo{}
\ReceiveDate{}
\ReviseDate{}
\AcceptDate{}
\OnlineDate{}
\AuthorMark{}
\AuthorCitation{Tian J M, Li L Y, Ye W T, et al.}

\title{Toward Real-World Table Agents: Capabilities, Workflows, and Design Principles for LLM-based Table Intelligence}{Tian J M, Li L Y, Ye W T, et al. Toward Real-World Table Agents: Capabilities, Workflows, and Design Principles for LLM-based Table Intelligence}


\author[1]{Jiaming TIAN}{}
\author[1]{Liyao LI}{}
\author[1]{Wentao YE}{}
\author[2]{Haobo WANG}{}
\author[1]{Lingxin WANG}{}
\author[3]{\\ Lihua YU}{}
\author[1,4]{Zujie REN}{{renzju@zju.edu.cn}}
\author[1]{Gang CHEN}{}
\author[1]{Junbo ZHAO}{}


\address[1]{College of Computer Science and Technology, Zhejiang University, Hangzhou 310007, China}
\address[2]{College of Software Technology, Zhejiang University, Ningbo 315048, China}
\address[3]{Bank Of HangZhou, Hangzhou 310016, China}
\address[4]{Zhejiang Lab, Hangzhou 311121, China}

\abstract{Tables are fundamental in domains such as finance, healthcare, and public administration, yet real-world table tasks often involve noise, structural heterogeneity, and semantic complexity—issues underexplored in existing research that primarily targets clean academic datasets. This survey focuses on \textbf{LLM-based Table Agents}, which aim to automate table-centric workflows by integrating preprocessing, reasoning, and domain adaptation. We define five core competencies—\textbf{C1:} Table Structure Understanding, \textbf{C2:} Table and Query Semantic Understanding, \textbf{C3:} Table Retrieval and Compression, \textbf{C4:} Executable Reasoning with Traceability, and \textbf{C5:} Cross-Domain Generalization—to analyze and compare current approaches. In addition, a detailed examination of the Text-to-SQL Agent reveals a performance gap between academic benchmarks and real-world scenarios, especially for open-source models. Finally, we provide actionable insights to improve the robustness, generalization, and efficiency of LLM-based Table Agents in practical settings.}


\maketitle

\section{Introduction}
Tables are ubiquitous in daily life and extensively utilized across diverse domains, including finance \cite{finance-table}, healthcare \cite{EHRAgent}, public administration \cite{public-administration-table}, and chemistry \cite{LLM-then-Xgboost}, playing a pivotal role in modern society.
Despite the remarkable progress of Large Language Models (LLMs) like GPT-4 \cite{GPT-4} in table tasks, most research remains limited to academic datasets such as Spider \cite{Spider} and WikiTQ \cite{wikitq}, which exhibit clean structure and simplified semantics. In contrast, real-world tables pose significant challenges due to noise, structural heterogeneity, and semantic ambiguity.
To address this gap, a new generation of intelligent systems—\textbf{LLM-based Table Agents} like SheetAgent \cite{SheetAgent} and ReAcTable \cite{ReAcTable}—has emerged. 
As a specialized branch of LLM-based Agents \cite{survey-agent, survey-agent-fcs, survey-agent-xi}, these systems aim to manage end-to-end table workflows by autonomously integrating table preprocessing, reasoning, and domain adaptation.

This survey diverges from prior modular reviews by focusing on the core capabilities essential for real-world table intelligence. 
We first identify five critical competencies required for LLM-based Table Agents: \textbf{C1. Table Structure Understanding}, which involves formatting tables for LLM input and handling complex features like merged cells and hierarchical headers; \textbf{C2. Table and Query Semantic Understanding}, addressing noisy or ambiguous data and improving query-table alignment; \textbf{C3. Table Retrieval and Compression}, which entails compressing or selecting large tables while preserving semantics; \textbf{C4. Executable Reasoning with Traceability}, ensuring agents generate verifiable intermediate steps; and \textbf{C5. Cross-Domain Generalization}, enabling rapid adaptation to different domains, such as finance and healthcare.
Next, we examine current research from a capability-centric perspective, highlighting key methods and their limitations. Finally, we outline a roadmap and actionable steps for developing next-generation LLM-based Table Agents that are safe, efficient, and generalizable.

\definecolor{inferenceColor}{HTML}{f95738}  
\definecolor{sftColor}{HTML}{ee964b}        
\definecolor{rlColor}{HTML}{f4d35e}          
\definecolor{pretrainColor}{HTML}{faf0ca}    
\definecolor{futureColor}{HTML}{bdd5ea}      

\tikzstyle{my-box}=[
    rectangle,
    draw=hidden-black,
    rounded corners,
    text opacity=1,
    minimum height=1.5em,
    minimum width=5em,
    inner sep=2pt,
    align=center,
    fill opacity=.5,
]
\tikzstyle{leaf}=[
    my-box, 
    minimum height=1.5em,
    fill=hidden-green!50, 
    text=black,
    align=left,
    font=\normalsize,
    inner xsep=2pt,
    inner ysep=2pt,
]

\tikzstyle{inference}=[leaf, fill=inferenceColor!50]
\tikzstyle{sft}=[leaf, fill=sftColor!50]
\tikzstyle{rl}=[leaf, fill=rlColor!50]
\tikzstyle{pretrain}=[leaf, fill=pretrainColor!50]
\tikzstyle{future}=[leaf, fill=futureColor!50]

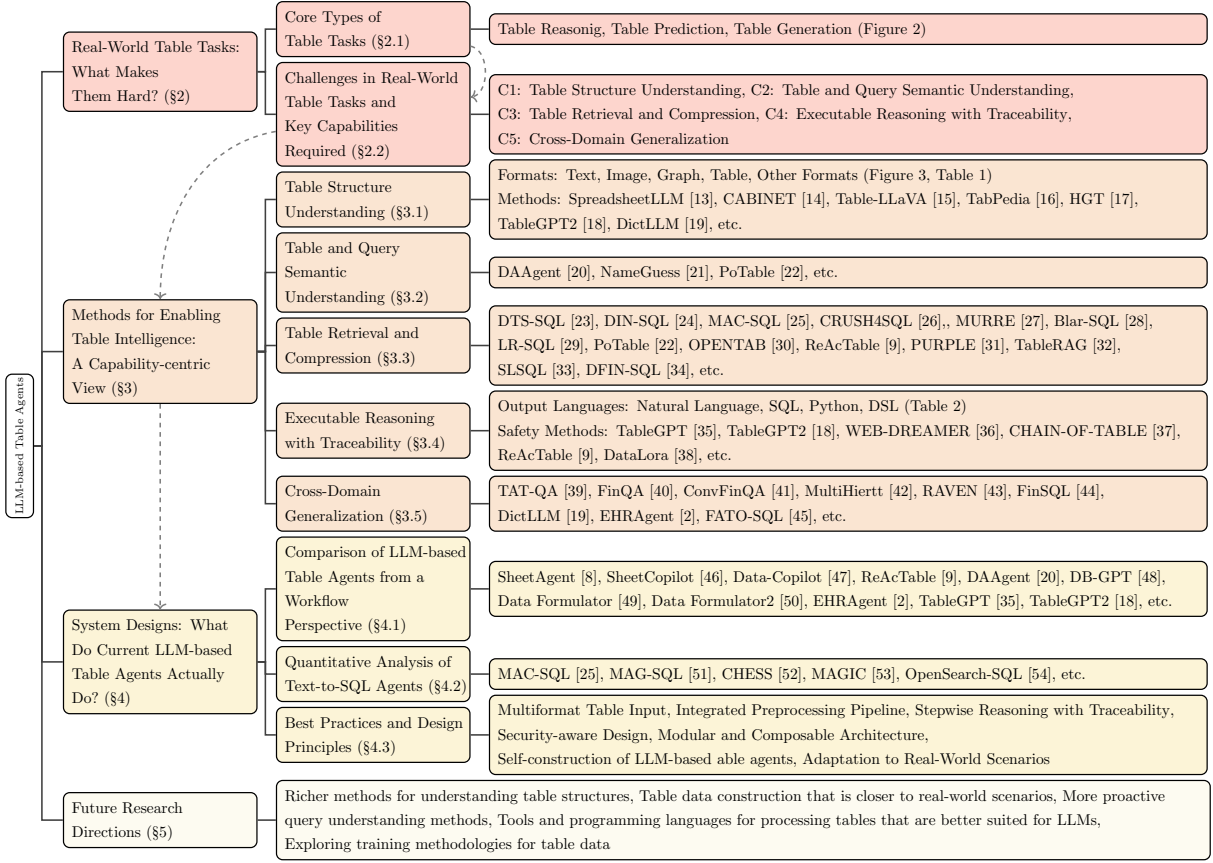
\begin{figure*}[!t]
    \vspace{-2mm}
    \centering
    \resizebox{\textwidth}{!}{
        \begin{forest}
            forked edges,
            for tree={
                child anchor=west,
                parent anchor=east,
                grow'=east,
                anchor=west,
                base=left,
                font=\footnotesize,
                rectangle,
                draw=hidden-black,
                rounded corners,
                align=left,
                minimum width=4em,
                edge+={darkgray, line width=1pt},
                s sep=3pt,
                inner xsep=2pt,
                inner ysep=3pt,
                line width=0.8pt,
                ver/.style={rotate=90, child anchor=north, parent anchor=south, anchor=center},
            },
            where level=1{text width=12em,font=\footnotesize,}{},
            where level=2{text width=12em,font=\footnotesize,}{},
            where level=3{text width=12em,font=\footnotesize,}{},
            where level=4{text width=10em,font=\footnotesize,}{},
            [
                LLM-based Table Agents, ver 
                [
                    ~Real-World Table Tasks: \\ ~What Makes \\ ~Them Hard?~(\S\ref{sec:real_world_task}), inference, name=realworld
                    [
                        ~Core Types of \\ ~Table Tasks~(\S\ref{sec:table_type}), inference, name=coretypes
                        [
                            ~Table Reasonig{,} 
                            Table Prediction{,}  
                            Table Generation (Figure \ref{fig:table-tasks}), inference, text width=45.6em 
                        ]
                    ]
                    [
                        ~Challenges in Real-World \\ ~Table Tasks and \\ ~Key Capabilities \\ ~Required~(\S\ref{sec:c1_5}), inference, name=challenges
                        [
                            ~C1: Table Structure Understanding{,} 
                            C2: Table and Query Semantic Understanding{,} 
                            \\
                            ~C3: Table Retrieval and Compression{,} 
                            C4: Executable Reasoning with Traceability{,}\\
                            ~C5: Cross-Domain Generalization, inference, text width=45.6em 
                        ]
                    ]
                ]
                [
                    ~Methods for Enabling \\ ~Table Intelligence: \\ ~A Capability-centric \\ ~View~(\S\ref{sec:c1_5_methods}), sft, name=methods
                    [
                        ~Table Structure \\ ~Understanding~(\S\ref{sec:c1}), sft
                        [
                            ~Formats: Text{,}
                            Image{,}
                            Graph{,}
                            Table{,}
                            Other Formats (Figure \ref{fig:modal}{,} Table \ref{tab:format_comparison})
                            \\ ~Methods: SpreadsheetLLM \cite{SpreadsheetLLM}{,}
                            CABINET  \cite{CABINET}{,}
                            Table-LLaVA  \cite{Table-LLaVA}{,}
                            TabPedia  \cite{TabPedia}{,}
                            HGT  \cite{HGT}{,}
                            \\ ~TableGPT2  \cite{TableGPT2}{,}
                            DictLLM  \cite{DictLLM}{, etc.}
                            , sft, text width=45.6em 
                        ]
                    ]
                    [
                        ~Table and Query \\ ~Semantic \\ ~Understanding~(\S\ref{sec:c2}), sft
                        [
                            ~DAAgent~\cite{DAAgent}{,}
                            NameGuess~\cite{NameGuess}{,}
                            PoTable~\cite{PoTable}{, etc.}
                            , sft, text width=45.6em 
                        ]
                    ]
                    [
                        ~Table Retrieval and \\ ~Compression~(\S\ref{sec:c3}), sft
                        [
                            ~DTS-SQL~\cite{DTS-SQL}{,}
                            DIN-SQL~\cite{DIN-SQL}{,}
                            MAC-SQL~\cite{MAC-SQL}{,}
                            CRUSH4SQL~\cite{CRUSH4SQL}{,}{,}
                            MURRE~\cite{MURRE}{,}
                            Blar-SQL~\cite{Blar-SQL}{,}
                            \\ ~LR-SQL~\cite{LR-SQL}{,}
                            PoTable~\cite{PoTable}{,}
                            OPENTAB~\cite{OpenTab}{,}
                            ReAcTable~\cite{ReAcTable}{,}
                            PURPLE~\cite{PURPLE}{,}
                            TableRAG~\cite{TableRAG}{,}
                            \\ ~SLSQL~\cite{SLSQL}{,}
                            DFIN-SQL~\cite{DFIN-SQL}{, etc.}
                            , sft, text width=45.6em
                        ]
                    ]
                    [
                        ~Executable Reasoning \\ ~with Traceability~(\S\ref{sec:c4}), sft
                        [
                            ~Output Languages: Natural Language{,}
                            SQL{,}
                            Python{,}
                            DSL (Table \ref{tab:language_comparison})
                            \\ ~Safety Methods: TableGPT~\cite{TableGPT}{,}
                            TableGPT2~\cite{TableGPT2}{,}
                            WEB-DREAMER~\cite{WEB-DREAMER}{,}
                            CHAIN-OF-TABLE~\cite{CHAIN-OF-TABLE}{,}
                            \\ ~ReAcTable~\cite{ReAcTable}{,}
                            DataLora~\cite{DataLora}{, etc.}
                            , sft, text width=45.6em 
                        ]
                    ]
                    [
                        ~Cross-Domain \\ ~Generalization~(\S\ref{sec:c5}), sft
                        [
                            ~TAT-QA~\cite{TAT-QA}{,}
                            FinQA~\cite{FinQA}{,}
                            ConvFinQA~\cite{Convfinqa}{,}
                            MultiHiertt~\cite{MultiHiertt}{,}
                            RAVEN~\cite{RAVEN}{,}
                            FinSQL~\cite{FinSQL}{,}
                            \\ ~DictLLM~\cite{DictLLM}{,}
                            EHRAgent~\cite{EHRAgent}{,}
                            FATO-SQL~\cite{FATO-SQL}{, etc.}
                            , sft, text width=45.6em  
                        ]
                    ]
                ]
                [
                    ~System Designs: What \\ ~Do Current LLM-based \\ ~Table Agents Actually \\ ~Do?~(\S\ref{sec:exp}), rl, name=system
                    [
                        ~Comparison of LLM-based \\ ~Table Agents from a \\ ~Workflow \\ ~Perspective~(\S\ref{sec:exp1}), rl
                        [
                            ~SheetAgent \cite{SheetAgent}{,}
                            SheetCopilot \cite{SheetCopilot}{,}
                            Data-Copilot \cite{Data-copilot}{,}
                            ReAcTable \cite{ReAcTable}{,}
                            DAAgent \cite{DAAgent}{,}
                            DB-GPT \cite{DB-GPT}{,}
                            \\ ~Data Formulator \cite{Data-Formulator}{,}
                            Data Formulator2 \cite{Data-Formulator-2}{,}
                            EHRAgent \cite{EHRAgent}{,}
                            TableGPT \cite{TableGPT}{,}
                            TableGPT2 \cite{TableGPT2}{, etc.}
                            , rl, text width=45.6em 
                        ]
                    ]
                    [
                        ~Quantitative Analysis of \\ ~Text-to-SQL Agents~(\S\ref{sec:exp2}), rl
                        [
                            ~MAC-SQL \cite{MAC-SQL}{,}
                            MAG-SQL \cite{MAG-SQL}{,}
                            CHESS \cite{CHESS}{,}
                            MAGIC \cite{MAGIC}{,}
                            OpenSearch-SQL \cite{OpenSearch-SQL}{, etc.}
                            , rl, text width=45.6em 
                        ]
                    ]
                    [
                        ~Best Practices and Design \\ ~Principles~(\S\ref{sec:result}), rl
                        [
                            ~Multiformat Table Input{,}
                            Integrated Preprocessing Pipeline{,}
                            Stepwise Reasoning with Traceability{,}
                            \\ ~Security-aware Design{,}
                            Modular and Composable Architecture{,}
                            \\ ~Self-construction of LLM-based able agents{,}
                            Adaptation to Real-World Scenarios
                            , rl, text width=45.6em 
                        ]
                    ]
                ]
                [
                    ~Future Research \\ ~Directions~(\S\ref{sec:future}), pretrain
                    [
                        ~Richer methods for understanding table structures{,}
                        Table data construction that is closer to real-world scenarios{,}
                        More proactive \\ ~query understanding methods{,}
                        Tools and programming languages for processing tables that are better suited for LLMs{,}
                        \\ ~Exploring training methodologies for table data
                        , pretrain, text width=59.45em 
                    ]
                ]
            ]
            \draw[dashed, ->, gray, line width=1pt] (coretypes) to[out=-10,in=10] (challenges);
            \draw[dashed, ->, gray, line width=1pt] (challenges) to[out=190,in=north] (methods);
            \draw[dashed, ->, gray, line width=1pt] (methods) to[out=south,in=north] (system);
        \end{forest}
    }
    \caption{Architecture of LLM-based Table Agents: existing methods, design principles, and future research directions. Dashed arrows denote conceptual progression, where the analysis of one module prompts the introduction of another.}
    \label{fig:taxonomy}
\end{figure*}

By shifting the focus from \textbf{``what exists''} to \textbf{``what is needed''}, this survey provides both a comprehensive synthesis of the field and a forward-looking blueprint for advancing table intelligence research.

The paper’s main structure is shown in Figure \ref{fig:taxonomy}.
The contributions of this paper are as follows:

\begin{enumerate}
    \item We present a novel perspective on surveying the field of \textbf{LLM-based Table Agents}, shifting the focus from existing solutions to the core capabilities required for real-world table intelligence. 
    
    \item We provide a detailed, capability-centric analysis of current methods, systematically summarizing existing approaches and highlighting their strengths and limitations in handling table tasks. 
    
    \item We conduct a \textbf{qualitative analysis} of several LLM-based Table Agents, examining how these systems integrate table preprocessing, reasoning, and domain adaptation within their workflows to effectively manage real-world table tasks.
    
    \item We perform a \textbf{quantitative analysis} of a specific variant, the \textbf{Text-to-SQL Agents}, demonstrating that most methods applied to academic datasets have little to no effect or only yield marginal improvements on weaker open-source models, thereby highlighting the challenges of applying these methods outside of idealized scenarios. 
    
    \item We offer actionable recommendations for future research directions, emphasizing the need for advancements in agent generalization, robustness, and efficiency to improve the applicability of LLM-based Table Agents in real-world scenarios across diverse domains and datasets. 
\end{enumerate}

\section{Real-World Table Tasks: What Makes Them Hard?}
\label{sec:real_world_task}
\subsection{Core Types of Table Tasks and LLM Focus Areas}
\label{sec:table_type}

Core Table tasks fall into three main types: table reasoning, table prediction, and table generation:

\begin{figure*}[htbp]
    \centering
    \includegraphics[width=\textwidth]{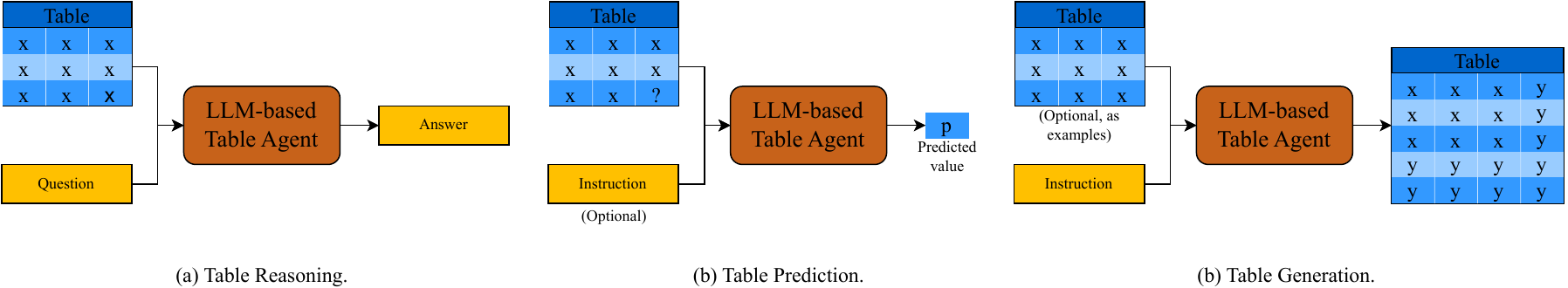}
    \caption{Examples of input and output for three types of table tasks—table reasoning, table prediction, and table generation. In the figure, "x" represents the known table content, "?" indicates the content to be predicted, "p" denotes the predicted value for "?" by the LLM-based Table Agent, and "y" represents the newly generated table content (note that table generation may not have any known table content).}
    \label{fig:table-tasks}
\end{figure*}

\begin{enumerate}
    \item \textbf{Table Reasoning.}
    Various papers offer different definitions of table reasoning \cite{Table-Reasoning-Survey, table-reasoning-definition}. 
    Here, we collectively refer to tasks where LLMs answer user queries based on a given table as table reasoning. 
    Tasks in table reasoning typically include \textbf{Table Question Answering (TableQA)} \cite{TableQA-definition}, \textbf{Table Fact Verification} \cite{Tabfact-definition}, \textbf{Table-to-Text} \cite{Table-to-Text-definition}, \textbf{Table Interpretation} \cite{Table-interpretation-definition}, etc.
    Table reasoning is the most complex task and the one most commonly addressed by current LLM-based Table Agents  \cite{SheetAgent, TableGPT, DAAgent, ReAcTable}. 
    It involves understanding table information to answer questions or perform actions on the table.

    \item \textbf{Table Prediction.} 
    Predicts missing table values based on available data, typically producing outputs of relatively fixed types.

    \item \textbf{Table Generation.} 
    Involves generating tables, whether in serialized form or through the creation of code or instructions that directly generate or guide table creation.
\end{enumerate}

Brief examples of these three tasks are illustrated in Figure \ref{fig:table-tasks}.
In current LLM-related academic research, table reasoning constitutes the majority, while studies on table prediction and table generation remain relatively scarce. 
In real-world applications, table reasoning is also where LLMs are most critically needed. 
Therefore, this survey primarily focuses on research related to table reasoning.

\subsection{Challenges in Real-World Table Tasks and Key Capabilities Required}
\label{sec:c1_5}

In academic datasets, table tasks have primarily focused on \textbf{Text-to-SQL} \cite{Spider}, \textbf{simple TableQA} \cite{wikitq}, and \textbf{Table Fact Verification} \cite{Tabfact-definition}. 
While there also exist datasets targeting complex or multi-hop reasoning \cite{Hybridqa}, the tables involved in those tasks tend to be relatively simple.
Real-world applications, however, demand far greater complexity. For example, analyzing financial reports may require combining Text-to-SQL with data visualization \cite{Data-Formulator, Data-Formulator-2} to extract \textbf{insights} \cite{DataLab}. Conversely, summarizing analytical reports into structured tables \cite{DocTabQA} presents an inverse challenge.

On the other hand, many earlier academic datasets feature small tables. 
Even in datasets with heterogeneous tables such as HiTab \cite{HiTab}, the complexity remains insufficient due to the limited size of the tables themselves. 
Only recently have datasets more suitable for LLMs—such as BIRD \cite{BIRD} and TQA-Bench \cite{TQA-Bench}—begun to include very long and complex tables. 
However, the task diversity and table types they cover are still quite limited. 
Current mainstream models, such as LLaMA 3 \cite{llama-3} and Qwen 3 \cite{qwen3}, typically support up to 128K context length. 
A few models can handle 1M \cite{qwen25-1M} or even 10M \cite{llama4} tokens, but this is still far from sufficient for many real-world scenarios. 

To overcome the limitations of academic datasets and enable LLM-based Table Agents to handle real-world scenarios effectively, we believe the following capabilities are essential:

\begin{enumerate}
\item \textbf{C1: Table Structure Understanding.}
The first step is to \textbf{enable LLMs to understand tables}—i.e., determining the optimal format for inputting tables into the model such that structural information is preserved. Tables vary widely, from well-structured relational database tables \cite{DB-table} and CSVs to complex tables with hierarchical headers \cite{HiTab} and additional elements \cite{SheetAgent}.

A more advanced requirement is the ability to handle merged cells, metadata (e.g., highlights), hierarchical headers, and layout irregularities. 
Tables also exhibit characteristics such as \textbf{permutation invariance}, where table semantics remain unchanged if rows or columns are reordered. Ideally, the model's output should be invariant to such transformations \cite{Permutation-invariance}, but the use of position encoding schemes such as RoPE \cite{RoPE} in LLMs makes this challenging. Addressing such structural properties remains an important direction for Table Agent development.

\item \textbf{C2: Table and Query Semantic Understanding.}
Real-world tables are often noisy and semantically ambiguous due to issues like abbreviations and anonymization. 
Additionally, table data is \textbf{heterogeneous}, containing both numerical and categorical fields \cite{Deep-table-survey}, which further complicates semantic understanding. 
Effective preprocessing is often required to mitigate these issues.
Moreover, table tasks involve not only interpreting tables but also understanding user queries. 
Accurately interpreting queries in context and resolving ambiguities are critical capabilities.

\item \textbf{C3: Table Retrieval and Compression.}
Many tables are \textbf{sparse}. 
Although large in size, only a small subset of data is usually relevant. 
Given the limited context length of current LLMs, even if models are technically capable of processing very large tables, doing so can lead to increased computational cost and performance degradation \cite{Challenges-of-LLM}.
Although many tasks can be accomplished using only the schema and SQL, tasks like TAG \cite{TAG} demonstrate that there are still numerous tasks where the table content must be input into the LLM, and programming languages such as SQL and Python cannot handle all types of tables.
Therefore, large tables must be compressed or selectively retrieved in a way that preserves semantic fidelity.

\item \textbf{C4: Executable Reasoning with Traceability.}
A core question is: \textbf{\textit{how to effectively complete table tasks}}—specifically, in what form (e.g., natural language or code) the model should output information to achieve the desired goal.
Beyond producing final answers, LLM-based Table Agents must be able to generate safe, verifiable intermediate code or logical reasoning chains to ensure transparency and reliability.

\item \textbf{C5: Cross-Domain Generalization.}
Agents should be capable of adapting to new domains (e.g., healthcare, finance) with as little overhead as possible, aiming to sustain strong performance even in unfamiliar contexts.

\end{enumerate}

\section{Methods for Enabling Table Intelligence: A Capability-centric View}
\label{sec:c1_5_methods}
In this section, we systematically review prior work through the lens of the five identified capabilities.

\subsection{Table Structure Understanding}
\label{sec:c1}

\begin{figure}[htbp]
    \centering
    \includegraphics[width=\textwidth]{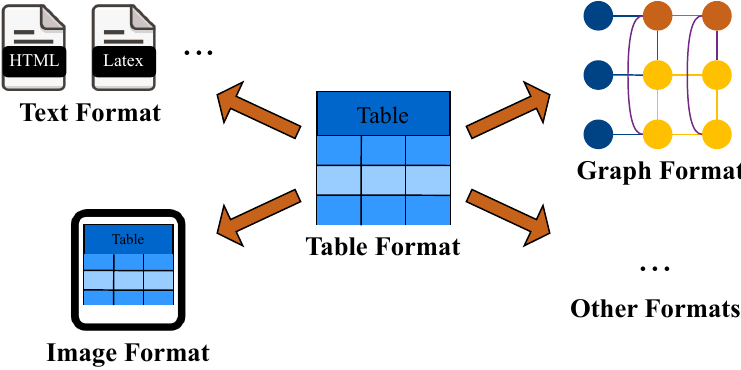}
    \caption{Various formats of table data for table reading.}
    \label{fig:modal}
\end{figure}

The table, as a special modality, presents a challenge when it comes to inputting it into LLMs. 
To read a table, it is first necessary to determine the format in which the table will be interpreted. 
The formats include \textbf{text}, \textbf{image}, \textbf{graph}, \textbf{table}, and \textbf{others}. 
Examples of these formats are illustrated in Figure \ref{fig:modal}.
For different formats, we aim to summarize their applicable scenarios, strengths, and weaknesses based on existing research, including their suitability for handling table data. 
This will help LLM-based Table Agents select the most appropriate format for table reading.

\subsubsection{Text Format}
One approach is to convert it into a \textbf{text format}, which is currently the most commonly used format. 
\textbf{Database tables}, \textbf{spreadsheets}, and other similar formats can relatively easily be transformed into text-based representations, with numerical and textual data types also being converted into text form. 
However, since tables are generally two-dimensional or even higher-dimensional structures, and LLMs process data as sequential input, it is necessary to serialize the table into a one-dimensional sequence.

Various serialization methods exist for tables, with \textbf{Markdown} format being the most common \cite{LLM-table-big-survey}. 
Markdown offers readability and requires fewer tokens. Other methods include \textbf{JSON} \cite{HTML-format}, \textbf{DFLoader} \cite{JSON-DFloader-format}, \textbf{Attribute-Value Pairs} \cite{Key-value-format}, \textbf{HTML} \cite{html-better}, \textbf{Latex} \cite{Latex-better}, and converting tables into natural language \cite{sentences-format-1, sentences-format-2}. 
HTML format has been found beneficial for GPT models due to their pre-training on web data \cite{html-better, HTML-format}. 
Latex has also shown promise \cite{Latex-better}. 
HTML and Latex may retain more structural information, aiding LLMs' understanding, but require more tokens. 
To further optimize performance, the FLEXTAF \cite{FLEXTAF} method dynamically selects the most suitable serialization method for specific tasks or integrates multiple serialization approaches.
Adding identifiers and highlighting key information in serialization can also impact LLMs' understanding of tables \cite{many-type-tables}.

In addition to the common table formats, some studies have attempted to design custom serialization methods that are more conducive to LLMs' understanding of tables. 
For example, SpreadsheetLLM \cite{SpreadsheetLLM} introduces a spreadsheet compression serialization technique to save input tokens.
CABINET \cite{CABINET} does not directly compress tables, it achieves denoising by assigning weights to each token.

Text format adapts more easily to LLMs, and many LLM training datasets already include substantial amounts of table data. 
However, LLMs face significant challenges with \textbf{permutation invariance}. 
Studies have shown that LLMs are highly sensitive to row-column permutations and transpositions in input tables \cite{LLM-bad-in-table-shuffle}. 
Additionally, this approach is constrained by the context length of LLMs, making it difficult to handle large tables effectively. 
Most critically, serializing a table inherently destroys its structural information. 
Even with infinite context length, this fundamental problem cannot be fully resolved.

\subsubsection{Image Format}
Another common table format is the \textbf{image format}, which includes tables that inherently exist in image form—such as tables extracted directly from documents in visual form—and tables rendered from other formats.
Since the emergence of \textbf{MLLMs(Multi-modal LLMs)} such as GPT-4V \cite{GPT-4V} and Gemini \cite{Gemini}, several investigations have aimed to evaluate MLLMs' comprehension of table or chart images \cite{GPT-4V-test}. 
Some MLLMs, like DeepSeek-VL \cite{DeepSeek-VL}, have even incorporated image-formatted table data during their pre-training phases.
Additionally, multimodal datasets like CMMU \cite{CMMU} and M-Paper \cite{M-Paper} contain image table data, while TableVQA-Bench \cite{TableVQA-Bench} is specifically designed for image-based table tasks.
There are now specialized MLLMs for handling image tables, such as Table-LLaVA \cite{Table-LLaVA} and TabPedia \cite{TabPedia}.

Concerning image tables, their visual attributes play a critical role in facilitating machine understanding. Deng et al. \cite{many-type-tables} demonstrated that applying different colors to individual rows can enhance table reasoning for multimodal large language models (MLLMs), although the observed performance gains are relatively limited for already strong LLMs. Their findings underscore the importance of visual cues, such as row highlighting, in supporting table-based reasoning. Complementary to this, studies like TableVQA-Bench \cite{TableVQA-Bench} adopted a more granular approach by differentiating visual components—including table structure, cell content, borders, and textual elements—during table rendering. However, these efforts did not explicitly investigate how such visual distinctions influence MLLMs’ comprehension and reasoning capabilities.

The advantage of using the image format is that research on vision-language models is relatively mature, and it allows for easy presentation of visual elements, such as highlights, in tables in their original format. 
The image format can support various types of \textbf{heterogeneous} tables, including even handwritten or irregular ones. 
Some studies \cite{many-type-tables} have shown that, for certain tasks, vision-language models perform comparably when handling both image and text tables.

However, there are clear challenges. 
The most significant issue remains the handling of large tables. 
Existing methods struggle to support extremely high-resolution images for such tables. 
Current research \cite{Table-LLaVA, TabPedia} primarily focuses on tables commonly found in academic papers and similar contexts. 
Additionally, a key limitation is that the visual format cannot effectively address the \textbf{permutation invariance} problem. 
For tables that need to be converted from other formats to the image format, the rendering process itself is still an open research question. 
Furthermore, when the table image lacks a corresponding text format, interacting with and modifying the table—such as through SQL or Python operations—becomes significantly more challenging.

\subsubsection{Graph Format}

Another possible approach is to transform tables into a \textbf{graph format}. 
For example, HGT \cite{HGT} converts tables into graphs and uses a graph encoder to process either the table or a global representation of the table, ultimately solving real-world table tasks through a LLM.

One advantage of the graph format is its \textbf{flexibility}. 
It can address the issue of \textbf{heterogeneous} table representations, and RDL \cite{RDL} even attempts to model entire databases using graphs, although this modeling approach has not yet been applied to LLMs. 
Furthermore, adopting a cross-linked list approach for table modeling can theoretically enhance the efficiency of handling sparse tables, substantially reducing the number of nodes required for processing.
The challenge of \textbf{permutation invariance} is another problem that many graph neural networks aim to solve, with a considerable body of research already dedicated to it \cite{Graph-Permutation-Invariance-1, Graph-Permutation-Invariance-2, Graph-Permutation-Invariance-3}. 
In summary, the graph format shows significant potential for better handling table structures.

However, there is currently limited research on the graph format. 
Many questions remain unanswered, such as whether tables processed by graph networks can truly perform well across a wide range of table tasks, which graph modeling approach is most effective, and how to choose the right graph neural network. 
These issues still require substantial experimentation.

\subsubsection{Table Format}

Although there has been considerable research on table neural networks and table pretraining \cite{TAPAS, TABERT, TURL, TUTA, TransTab, CT-BERT}, studies that directly treat tables as a distinct format are relatively limited due to the difficulty in seamlessly integrating them with LLMs. 
TableGPT2 \cite{TableGPT2} attempts to address this by using a table-specific Encoder, which generates column vector representations through cell embeddings, row and column attention, and column vector aggregation. 
These representations are then provided to the LLM, helping it better understand the column distribution and semantic information in the table. 
However, while the encoder resolves the issue of \textbf{permutation invariance}, the order of column vectors still affects the model's output. 
Additionally, this Table Encoder cannot handle \textbf{heterogeneous tables}, making it more of an initial exploration in this area.

\subsubsection{Other Formats}
In addition to the common table-related formats, other studies have explored alternative approaches. 
For instance, DictLLM \cite{DictLLM} directly combines column names and cell meanings into a dictionary set, treating the table as a dictionary format. 
However, this method is typically limited to domain-specific tables and requires further investigation to evaluate its broader applicability.

\subsubsection{Comparison between Different Formats}

A simple comparison of these formats is presented in Figure \ref{tab:format_comparison}. 
Currently, the text format remains the most extensively studied due to its ease of adaptation to LLMs. While research on image format is growing, other formats have received relatively limited attention.
From a theoretical perspective, the table format best preserves the inherent characteristics of table data. 
However, in practice, the graph format has shown more advantages for table representation. 
It's worth noting that superior representational capability does not necessarily translate to better performance in downstream tasks.
Presently, comparative studies between these formats remain scarce \cite{many-type-tables}, making it difficult to determine which format will ultimately prove most effective. 
Clearly, no universally optimal format exists for table processing at this stage. Until more robust solutions emerge, researchers should focus on flexibly selecting formats based on specific requirements, balancing the needs of their systems—or even exploring hybrid approaches that combine multiple formats for enhanced performance.

\begin{table*}[htbp]
\centering
\centering
\caption{Comparison of different table formats. The contents reflect the current research stage, indicating whether these formats can address certain challenges in table data processing. Note that the limitations of these formats may be resolved with further advancements.}
\resizebox{1\linewidth}{!}{
\begin{tabular}{p{0.1\linewidth}
>{\centering\arraybackslash}p{0.12\linewidth}
>{\centering\arraybackslash}p{0.12\linewidth}
>{\centering\arraybackslash}p{0.12\linewidth}
>{\centering\arraybackslash}p{0.2\linewidth}
>{\centering\arraybackslash}p{0.3\linewidth}}
\toprule
\textbf{Formats} & \textbf{Research Focus in LLMs} & \textbf{Handles Permutation Invariance} & \textbf{Handles Heterogeneous Tables} & \textbf{Compresses Sparse Tables Efficiently} & \textbf{Other Characteristics} \\ \hline
\textbf{Text} & Highest & No & Yes & Yes (e.g., JSON format) & - \\ 
\textbf{Image} & Moderate & No & Yes & No & Easily handles tables with visual elements (e.g., handwritten tables) but is challenging for table editing. \\ 
\textbf{Graph} & Low & Yes & Yes & Yes & - \\ 
\textbf{Table} & Low & Yes & No & Yes & - \\ 
\textbf{Others} & Low & Yes & No & Yes & Limited research available; differences between formats may be significant. \\ \bottomrule
\end{tabular}
}
\label{tab:format_comparison}
\end{table*}

\subsection{Table and Query Semantic Understanding}
\label{sec:c2}
\subsubsection{Table Preprocessing}
Real-world tables are often characterized by \textbf{noise} and \textbf{semantic ambiguity}, commonly exhibiting \textbf{missing values} and \textbf{inconsistent formatting}. While methods such as DAAgent \cite{DAAgent} have integrated traditional preprocessing techniques into LLM-based agent workflows, can these approaches sufficiently improve the semantic comprehension of tables by LLMs? Prior studies indicate that semantic clarity and completeness, as opposed to compressed or ambiguous representations, play a critical role in influencing LLM performance. To this end, schema-level preprocessing becomes indispensable, encompassing tasks such as \textbf{column name normalization} and \textbf{schema construction}.

\paragraph{I. Column Name Cleaning}
\textbf{Column name cleaning} primarily focuses on refining column names to provide clear semantic information, which typically facilitates better table comprehension by LLMs. Traditionally, this task has been approached as a classification problem \cite{NameGuess-classification-problems-1, NameGuess-classification-problems-2}, where abbreviated terms are mapped to predefined full names.
However, NameGuess \cite{NameGuess} extends this by framing it as a generative task, using LLMs to reconstruct full column names. 
Despite its advantages, NameGuess has limitations, such as its inability to address omitted details in column names and its lack of consideration for column relationships or data distribution within the table. 
These shortcomings hinder its ability to fully understand the table's structure. 
One potential approach is to model the table data using neural networks and provide table embeddings to the LLM, allowing it to access more comprehensive information beyond just the schema.

\paragraph{II. Schema Construction and Column Relationship Detection}
\textbf{Schema construction and column relationship detection} can be used to convert tables into formats compatible with tools like SQL or Python, making them easier to process programmatically. 
Research in column type detection \cite{Column-type-detection-1, Column-type-detection-2, Column-type-detection-3} and column relationship identification \cite{column-relationship-identification-1, column-relationship-identification-2} has contributed to this area. 
For instance, PoTable \cite{PoTable} employs LLMs for data type cleaning, combining CoT reasoning with Python code execution to ensure accurate data types. 
However, existing methods, including PoTable, still do not leverage the full potential of LLMs' intrinsic knowledge in schema construction. 
This remains an important area for future exploration.

\subsubsection{User Query Understanding}

Text queries often pose challenges due to inherent \textbf{ambiguity}.
To address this, advanced \textbf{intent detection} within LLMs is crucial \cite{Intent-Detection}. 
When faced with vague or unclear queries, models must demonstrate robustness and the ability to follow instructions \cite{Instruction-following}. 
In cases of excessively ambiguous queries, it may be necessary for models to prompt users for clarification \cite{Code-clarify}. 
\textbf{Intention clarification} is a common strategy for dealing with such ambiguity. 
This can involve direct questioning or asking the user for more details. 
Several systems, like SheetAgent \cite{SheetAgent} and TableGPT \cite{TableGPT}, incorporate intent detection capabilities to better understand user queries. 
Multi-turn datasets like CoSQL \cite{Cosql} also require models to seek clarification from users, especially when handling ambiguous queries.

In table contexts, ambiguity presents additional complexities. Human annotators only agree 62\% of the time on whether a given SQL query is ambiguous \cite{floratouNL2SQLSolvedProblem2024}, highlighting the subjective nature of ambiguity. 
Despite this, progress has been made in modeling ambiguity in queries. 
MQS-SQL \cite{MQA-SQL} leverages the power of strong LLMs to reduce ambiguity. Papicchio et al. proposed a pipeline for evaluating a model’s ability to handle ambiguous inputs \cite{papicchioEvaluatingAmbiguousQuestionsa}, though several types of ambiguity remain unresolved. To advance this line of work, Hu et al. \cite{hu2025ambiguity} further explored ambiguity from the perspective of latent space, aiming to both analyze and alleviate it.

One promising approach to dealing with ambiguity is \textbf{top-k generation}, where multiple candidate answers are proposed for the user to select from \cite{bhaskarBenchmarkingImprovingTexttoSQL2023}. 
This allows models to handle ambiguous queries more effectively. 
Additionally, documentation for human users has been found to assist models like GPT-4 in Text-to-SQL tasks, enabling models to better navigate ambiguous data structures \cite{huangDataAmbiguityStrikes2023}. 
This suggests that external context, in addition to the query itself, plays a crucial role in resolving ambiguity in table-based queries.

\subsection{Table Retrieval and Compression}
\label{sec:c3}

A common method for \textbf{Table Retrieval and Compression} is \textbf{schema link}, a technique that identifies relevant tables, columns, and key values from a database schema to answer a user’s query \cite{DIN-SQL}. But how can we determine which parts of the schema are truly necessary for a given task? The simplest approach employs LLMs to directly predict schema links using the schema and query (e.g., DTS-SQL \cite{DTS-SQL}, DIN-SQL \cite{DIN-SQL}, MAC-SQL \cite{MAC-SQL}). For larger databases, a two-stage process is often used: first retrieving candidate tables, then performing column-level linking within them (e.g., CRUSH4SQL \cite{CRUSH4SQL}, MURRE \cite{MURRE}). When schemas exceed LLM context limits, methods like Blar-SQL \cite{Blar-SQL} and LR-SQL \cite{LR-SQL} propose schema chunking—splitting the schema into manageable segments and merging results. Future work could explore dynamic chunking strategies to improve efficiency.

However, schema link mainly suits relational databases. 
Other methods aim to \textbf{filter essential information} from diverse table types. 
For example, PoTable \cite{PoTable} implements row selection.
OPENTAB \cite{OpenTab} conducts column selection via SQL first, then proceeds with row and column selection. 
ReAcTable's \cite{ReAcTable} primary operation involves selecting rows and columns. 
PURPLE \cite{PURPLE} represents schema as a graph and uses the Steiner \cite{Steiner-tree} tree problem to prune the schema.
TableRAG \cite{TableRAG} takes table retrieval a step further by refining it to the cell level, effectively eliminating irrelevant data.

\textbf{Table Retrieval and Compression} directly influence the performance of downstream table-related tasks, as errors introduced at this stage can propagate and compromise the effectiveness of subsequent reasoning or generation. Despite its importance, this stage has received limited attention in terms of systematic evaluation. Existing methods, such as SLSQL \cite{SLSQL}, often rely on standard metrics like Precision, Recall, and F1-score, which may be insufficient to capture the impact of retrieval quality on downstream performance. To better accommodate LLMs' needs—particularly in schema linking—DFIN-SQL \cite{DFIN-SQL} introduces the \textbf{Schema Link Accuracy Metric}. However, there is still a lack of well-designed datasets and benchmarks to rigorously assess retrieval and compression quality.

\subsection{Executable Reasoning with Traceability}
\label{sec:c4}

\subsubsection{Code-based Table Reasoning}
The most straightforward approach to table tasks is to prompt LLMs to generate answers directly through \textbf{natural language} output. 
This method typically relies on prompt engineering \cite{table_prompt_en}. For instance, ToolWriter \cite{ToolWriter} directly uses LLMs to generate answers for simple TableQA tasks. 
Sui et al. adopt self-augmented prompting \cite{HTML-format}, where the LLM first forms an internal understanding of the table before producing the answer. 
Some approaches further enhance reasoning by incorporating Chain-of-Thought (CoT) prompting \cite{CoT}, as shown by Liu et al. \cite{LLM-bad-in-table-shuffle}.
Influenced by methodologies such as o1 \cite{OpenAI-o1} and DeepSeek-R1 \cite{Deepseek-r1}, a growing body of research like Table-R1 \cite{Table-R1} and Reasoning-Table \cite{Reasoning-Table} has emerged focusing on Test-Time Scaling (TTS) techniques specifically tailored for TableQA tasks.

Despite these efforts, research on using LLMs for direct table reasoning remains limited. 
LLMs are not only constrained in their reasoning capabilities but also highly sensitive to the formatting of table inputs. Liu et al. \cite{LLM-bad-in-table-shuffle} demonstrated that simply transposing a table or reordering rows and columns can significantly degrade performance. 
These issues stem from fundamental limitations in LLM architectures and are currently difficult to resolve.

Moreover, there is often a trade-off between the traceability and capability scope of LLM-based Table Agents. 
Achieving traceability typically requires strict constraints on the output format, yet different tasks often necessitate task-specific formatting. 
This tension leads to high maintenance overhead and reduced flexibility.

To address these challenges, one promising direction is to have LLMs generate code instead of direct answers. 
Code-based outputs support executable operations on tables, such as transformation and visualization, thereby improving both expressiveness and traceability. There are three common types of programming languages—\textbf{SQL}, \textbf{Python}, and \textbf{domain-specific language (DSL)}.

\paragraph{I. SQL}
Various studies explore \textbf{Text-to-SQL}, utilizing datasets like Spider \cite{Spider}, BIRD \cite{BIRD} and WikiSQL \cite{WikiSQL}. 
Methods like C3 \cite{C3} employ prompt engineering for Text-to-SQL in zero-shot mode, while Nan et al. \cite{few-shot-NL2SQL} utilize few-shot learning. SEA-SQL \cite{SEA-SQL} goes further by exploring semantic enhancements for database schemas. QDecomp+InterCOL \cite{QDecomp+InterCOL} uses CoT to tackle Text-to-SQL challenges. 
Approaches such as TableLLaMA \cite{TableLlama} and UnifiedSKG \cite{Unifiedskg} are based on fine-tuning.

Effective Text-to-SQL methods often split the task into multiple stages. 
One common approach involves schema link followed by SQL generation, as seen in Blar-SQL \cite{Blar-SQL} and DTS-SQL \cite{DTS-SQL}. 
Another method first generates the SQL structure or an intermediate representation of SQL, then its content, as demonstrated by ZeroNL2SQL \cite{ZeroNL2SQL}, SC-Prompt \cite{SC-Prompt}, and OpenSearch-SQL \cite{OpenSearch-SQL}.
SGU-SQL \cite{SGU-SQL} enhances linkage between user queries and databases, then uses syntax trees to guide SQL generation. 
PET-SQL \cite{PET-SQL} generates preliminary SQL, performs schema link based on it, and then finalizes the SQL. 
ReBoostSQL \cite{ReboostSQL} and DIN-SQL \cite{DIN-SQL} primarily address query rewriting, schema link, SQL generation, and self-correction. 
DFIN-SQL (\cite{DFIN-SQL}), an enhancement of DIN-SQL, focuses on producing concise table descriptions. 
PURPLE \cite{PURPLE} focuses on schema pruning, skeleton prediction, demonstration selection, and database adaptation to accommodate SQL rule variations among different databases.

An alternative approach is to refine existing SQL queries rather than generating them from scratch. 
For instance, MAGIC \cite{MAGIC} analyzes failed SQL cases in the training data, creates self-correction guidelines, and maintains a guideline repository. 
These guidelines are then applied to other datasets to fix problematic SQL queries.
In addition to leveraging historical data, SQLFixAgent \cite{SQLFixAgent} employs an agent-based rubber duck debugging method to verify generated SQL, increasing the likelihood of correctness.

TTS has also been extensively studied in the context of SQL, demonstrating its versatility across different structured data tasks.
For instance, SQL-o1 \cite{SQL-o1} employs Monte Carlo Tree Search (MCTS) \cite{MCTS-survey} to implement TTS, while Reasoning-SQL \cite{Reasoning-SQL} and SQL-R1 \cite{SQL-R1} follow the approach of DeepSeek-R1, leveraging reinforcement learning with rewards tied to SQL syntax or execution to enhance the model's reasoning capabilities in SQL tasks. 
However, these studies are still at an early stage and largely represent straightforward replications of reasoning-model training paradigms like o1 and DeepSeek-R1, without delving into training strategies that are more tailored to the characteristics of SQL.

Nevertheless, Text-to-SQL still faces several issues:
\begin{enumerate}
\item SQL syntax remains limited, mostly revolving around \texttt{SELECT} statements. 
\texttt{ALTER} and \texttt{UPDATE} queries are notably scarce. 
In the BIRD \cite{BIRD} dataset, among 12,751 reference SQL queries analyzed, only 36 instances of \texttt{LEFT JOIN} or \texttt{RIGHT JOIN} were found, with the majority being \texttt{INNER JOIN}.
\item Evaluation of SQL queries is inadequate. 
Key metrics like Exact Matching and Execution Accuracy are commonly used, as in Spider \cite{Spider}, yet additional metrics such as Valid Efficiency Score are seldom employed, except in datasets like BIRD.
\item Some non-English Text-to-SQL datasets, such as CSpider \cite{Cspider}, are translated and adapted from English benchmarks like Spider \cite{Spider}, often leading to mismatches between question phrasing and table or column names. Moreover, column names are sometimes only implicitly reflected in the question semantics \cite{Chinese-Text-to-SQL-industrial}, making the alignment between natural language and schema less direct. These issues highlight the inherent difficulty in constructing high-quality non-English benchmarks, where direct translation may fail to capture linguistic and contextual nuances across languages.
\end{enumerate}

\paragraph{II. Python}
There are few datasets specifically designed for Python-based table tasks.
Datasets like HumanEval \cite{HumanEval}, MBPP \cite{MBPP}, and DS-1000 \cite{DS-1000} include problems with pandas but often lack complexity and table input.
Datasets like SOFSET \cite{SOFSET}, which focus on real-world StackOverflow problems with table inputs, are scarce.

Python has strengths in visualizing data and processing non-database tables.
PoTable \cite{PoTable} primarily uses Python as the execution language, obtaining feedback from the Python execution results to guide the entire workflow. 
Similarly, TableGPT2 \cite{TableGPT2} leverages Python for visualization, with the visual output generating feedback through a vision-language model.

However, some studies show Python may not always be advantageous for table reasoning.
For example, Liu et al. found Direct Prompting outperformed Python \cite{NL2SQL-Python}, and ReBoostSQL found Text-to-Python less effective than Text-to-SQL \cite{ReboostSQL}.
This may be because SQL is designed specifically for databases, and handling tables is only a small part of Python's capabilities. 
Targeted enhancements using Python specialized for table processing could help address this issue.

\paragraph{III. DSL}
Some studies investigate the effectiveness of DSL for table tasks. 
For instance, ReBoostSQL \cite{ReboostSQL} designs encapsulated functions and demonstrates superior performance compared to SQL generation methods on advanced models like GPT-4. 
CHAIN-OF-TABLE \cite{CHAIN-OF-TABLE} employs custom atomic operations to process tables step-by-step following the CoT approach. 
TableGPT \cite{TableGPT} defines a DSL for table reasoning.

Table \ref{tab:language_comparison} highlights the differences among several languages commonly used for table reasoning tasks. 

\textbf{Natural language} is the most familiar to LLMs and has a broad range of applications. 
However, it lacks structure and is not specifically designed for handling tables, making it less effective for table reasoning tasks.
\textbf{SQL} is the most widely used language for table tasks. 
It benefits from a well-established set of optimization strategies, making it highly efficient for working with relational databases. 
Its strong pretraining presence ensures that LLMs can often handle SQL with minimal additional effort.
\textbf{Python}, while versatile and broadly applicable, relies heavily on libraries like pandas for handling table data. 
These libraries are best suited for sheet-based tables. 
For other types of tables, Python may require integrating additional languages like SQL or custom solutions, which increases complexity.
\textbf{DSL}, in contrast, are designed for atomic operations, offering significant advantages in traceability and explainability. 
However, as these languages are not typically included in pretraining data, additional efforts are needed to prepare data and train LLMs to work effectively with them.

Therefore, different programming languages cater to distinct use cases. \textbf{Natural language}, despite its intuitive familiarity and broad applicability, proves less suitable for table reasoning tasks due to its inherent lack of specificity. \textbf{SQL} demonstrates superior performance in relational database operations, leveraging its well-optimized query structure. \textbf{Python} offers unparalleled flexibility across diverse computational scenarios, particularly when integrated with specialized libraries. \textbf{DSL}, while requiring higher initial development costs, provides significant advantages in applications demanding enhanced traceability and explainability.

In real-world applications, a hybrid approach is often recommended, combining languages based on task-specific needs to balance precision, flexibility, and maintainability under real-world constraints.

\begin{table*}[htbp]
\centering
\centering
\caption{Performance comparison of several commonly used languages for table reasoning tasks. The "Pretraining Inclusion" column primarily reflects the level of post-training difficulty, which varies significantly across languages.}
\resizebox{1\linewidth}{!}{
\begin{tabular}{p{0.15\linewidth}
>{\centering\arraybackslash}p{0.2\linewidth}
>{\centering\arraybackslash}p{0.2\linewidth}
>{\centering\arraybackslash}p{0.2\linewidth}
>{\centering\arraybackslash}p{0.3\linewidth}}
\toprule
\textbf{Language} & \textbf{Pretraining Inclusion} & \textbf{Handled Table Types} & \textbf{Designed for Table Data Processing} & \textbf{Traceability \& Explainability} \\ \hline
\textbf{Natural Language} & Yes & All text-format tables & No & Poor (can generate atomic step explanations) \\ 
\textbf{SQL} & Yes & Database & Yes & Poor (especially for complex nested cases) \\ 
\textbf{Python} & Yes & Usually sheet-based tables & No & Poor (usually has some atomicity) \\ 
\textbf{DSL}  & 
No & Theoretically all tables & Yes &
Good (each statement typically corresponds to a specific step) \\ \bottomrule
\end{tabular}
}
\label{tab:language_comparison}
\end{table*}

\subsubsection{Verifiable and Safe Reasoning over Table Data}

Security and privacy are critical considerations in AI systems \cite{SCIS-safe-survey}, especially when dealing with structured data. In real-world scenarios, tables are commonly used to store and process information in domains such as healthcare, finance, and government—areas that typically involve stringent security and privacy requirements.
For instance, banking operations often demand high levels of data \textbf{security} and \textbf{traceability}.
The introduction of tasks like Text-to-SQL also brings potential risks, such as the ToxicSQL \cite{ToxicSQL} method, which attempts a backdoor attack on Text-to-SQL and demonstrates the inadequacy of existing defense mechanisms against such attacks.
This means that for table data, in addition to preventing the generation of \textbf{biased}, \textbf{discriminatory}, or \textbf{privacy-violating} content, it is crucial to consider data \textbf{integrity}, \textbf{reliability}, \textbf{confidentiality}, and \textbf{traceability} \cite{Security-survey}.
Of course, implementing security measures often comes with additional costs, which can result in decreased efficiency and performance. Finding a balance between performance and security is also an important consideration.

There is significant research on adversarial attacks on LLMs \cite{LLM-Adversarial-2, LLM-Adversarial-1}. 
However, there is less research focused on table data. 
The primary concerns regarding the security of table data are privacy and data safety. 
Kim et al. \cite{kim2025SQL-Injection-Fuzzing} investigated SQL Injection fuzzing and detection using LLMs; however, their study only demonstrated the feasibility of employing LLMs for SQL injection fuzz testing and did not apply it in real-world scenarios.
TableGPT \cite{TableGPT} emphasizes private deployment to address these concerns. 
Its improved version, TableGPT2 \cite{TableGPT2}, creates a code sandbox environment to run the code.
However, such methods require setting up an additional environment, and even after ensuring security, some tasks still need to be run in a real environment, which leads to a decrease in efficiency.
A feasible improvement could be inspired by WEB-DREAMER \cite{WEB-DREAMER}, where LLMs are used as world models to predict the potential outcomes of taking a specific action. 
This approach may help maintain efficiency while ensuring a certain level of privacy and data safety.
The security of table data is also linked to the traceability of data processing. 
One approach is using DSL to limit LLM functionalities, preventing them from generating malicious code. 
Examples include CHAIN-OF-TABLE \cite{CHAIN-OF-TABLE} and ReAcTable \cite{ReAcTable}. 
Another method involves post-hoc mitigation. 
For example, DataLore \cite{DataLora} is a method that, given a table $A$ and its enhanced counterpart $A'$, aims to infer a plausible sequence of transformation operations that could have led from $A$ to $A'$, thereby enabling post-hoc analysis of table enhancement processes.

However, enforcing traceability and designing effective DSL can constrain LLM capabilities, making it challenging to balance performance and security. Existing table datasets largely ignore security aspects, highlighting the need for dedicated benchmarks.

\subsection{Cross-Domain Generalization}
\label{sec:c5}
Table tasks in different domains show distinct characteristics.
Considering the significant variations in table data across different domains, a general LLM-based Table Agent may struggle to meet the demands of all domains. 
To ensure that an LLM-based Table Agent can truly address real-world scenarios, domain adaptation becomes an essential capability. 
With this in mind, we have reviewed research efforts related to domain adaptation for table data. 
Our ultimate goal is to answer a key question: \textbf{\textit{Is there currently a domain adaptation method that strikes a balance between performance and cost?}}

Handling domain-specific table tasks presents unique challenges due to varying table structures, specialized terminologies, and complex computational requirements across fields like finance and medicine.
In the \textbf{financial domain}, tasks often involve complex mathematical computations, such as statistical data analysis and risk assessment \cite{FinQA}.
In the \textbf{medical domain}, there is a wide variety of table types, including checklists summarizing numerous physical and chemical indicators, various forms of medical reports \cite{DictLLM, EHRAgent}, and coefficient tables with a large number of columns used in drug development \cite{Many-feature-in-biomedical}. 
Different types of tables should be handled with different approaches.
Moreover, each domain has its specific terminologies, which require special handling in table tasks.

Researchers have developed diverse approaches to tackle table tasks across various fields.
Research in the \textbf{finance domain} has been extensive.
Representative datasets for TableQA in this domain include TAT-QA \cite{TAT-QA}, FinQA \cite{FinQA}, ConvFinQA \cite{Convfinqa}, and Multihiertt \cite{MultiHiertt}.
Benchmarks such as KnowledgeMATH \cite{Knowledgemath} for TableQA and BULL \cite{FinSQL} for Text-to-SQL emphasize financial knowledge and numerical reasoning capabilities.
To enhance TableQA and Text-to-SQL performance in the financial sector, beyond conventional fine-tuning and prompt engineering techniques \cite{Hwang-paper}, LLMs also integrate tool-based approaches, including RAVEN \cite{RAVEN} for structured reasoning and FinSQL \cite{FinSQL} for output calibration.
In the \textbf{medical domain}, current research mainly focuses on data with relatively fixed formats, such as DictLLM \cite{DictLLM}, which helps process structured data like medical laboratory reports, and EHRAgent \cite{EHRAgent}, which handles electronic health records (EHRs).
In the \textbf{materials domain}, Do et al. \cite{LLM-then-Xgboost} serialize tables into text, then use LLMs for embedding before employing XGBoost for table prediction.
In the \textbf{petrochemical industry}, many table column names are abbreviations of domain-specific terms. 
FATO-SQL \cite{FATO-SQL} primarily uses Retrieval-Augmented Generation (RAG) to help LLMs understand column names, enabling quick domain adaptation.

Table tasks across different domains exhibit distinct characteristics.
Our analysis reveals that existing cross-domain generalization methods have notable limitations: some RAG-based approaches offer limited performance improvements while requiring well-constructed databases; others relying on domain-specific training data are more costly than RAG implementations; while certain customized solutions designed for particular data characteristics demonstrate severely constrained applicability.
At present, no universal method exists for domain adaptation.
Each domain requires a deep understanding of its unique attributes and the development of targeted approaches that effectively balance performance and cost.

\section{System Designs: What Do Current LLM-based Table Agents Actually Do?}
\label{sec:exp}

\subsection{Comparison of LLM-based Table Agents from a Workflow Perspective}
\label{sec:exp1}
We analyze representative LLM-based Table Agents in terms of workflow coverage and capability alignment. By integrating components like parsing, reasoning, and tool use, these agents offer a holistic view of table-aware design.
In this section, we systematically compare their architectural choices, implementation maturity, and coverage of functional capabilities.

\textbf{SheetAgent} \cite{SheetAgent} is a framework with multi-agents. 
It employs iterative task reasoning and reflection to manipulate spreadsheets precisely and autonomously.
\textbf{SheetCopilot} \cite{SheetCopilot} devises atomic operations as abstractions of spreadsheet functions. 
They also developed a task-planning framework for interaction with large language models.
\textbf{Data-Copilot} \cite{Data-copilot} is a code-centric data analysis agent. 
It executes queries, processes, and visualizes data based on human requests.
\textbf{ReAcTable} \cite{ReAcTable} specializes in TableQA problems. It uses SQL and Python code in its processing, incorporating features like voting.
\textbf{DAAgent} \cite{DAAgent} performs table tasks using Python, referencing the ReAct \cite{ReAct} framework.
\textbf{DB-GPT} \cite{DB-GPT} understands natural language queries and generates accurate SQL queries. 
It includes a Python library for the convenience of developers.
\textbf{Data Formulator} \cite{Data-Formulator} assists in visualizing table data and automates table preprocessing for visualization pipelines.
\textbf{Data Formulator 2} \cite{Data-Formulator-2} is an enhanced version of Data Formulator, featuring innovative improvements in UI interaction. 
It also supports iteration functionality, allowing users to view and trace previous designs. 
However, there have been no significant changes to the task types related to skill execution.
\textbf{TableGPT} \cite{TableGPT} utilizes a Table Encoder for comprehensive table understanding. 
It handles various operations such as question answering, data manipulation, visualization, analysis report generation, and automated prediction.
\textbf{TableGPT2} \cite{TableGPT2} is an improved version of TableGPT. 
Compared to ChatGPT, it adds support for table-based input, replaces DSL with SQL and Python, and optimizes the agent workflow. 
It uses a code sandbox environment to ensure secure execution and incorporates a visual-language model to provide feedback on Python visualizations. 
Overall, the process is more refined and mature.
\textbf{EHRAgent} \cite{EHRAgent} focuses on EHRs and autonomously executes complex clinical tasks, integrating medical knowledge during processing.

A comparison of these agents based on workflow is available in Table~\ref{Agents-compare}. In the table, the \textbf{Table and Query Semantic Understanding} capability primarily corresponds to the \textbf{Table Types \& Formats} in the workflow.
The \textbf{Executable Reasoning with Traceability} capability is primarily associated with \textbf{Output Language Types}, \textbf{Handled Table Tasks}, and \textbf{Table Data Safety}.

Through a comparative analysis of the workflow, we have made the following observations:

\begin{enumerate}
    \item \textbf{Dominance of text formats in current agents.}
    Most agents primarily rely on text format. 
    These agents are relatively simple and benefit from well-established research in prompt engineering. 
    While there has been some progress in visual-language models for processing table data, these models often face limitations in editing and manipulating tables, restricting their application scope. 
    As a result, LLM-based Table Agents that leverage visual formats are still rare. 
    In the future, such agents may find utility in areas like paper comprehension. 
    TableGPT2 introduces an innovative table format. 
    However, its Table Encoder has limited compatibility with certain table types, forcing reliance on text format for unsupported formats.

    \item \textbf{Insufficient preprocessing of table data.}
    A common challenge is the inadequate preprocessing of table data. 
    Real-world tables are often noisy, yet many agents fail to leverage the inherent abilities of LLMs for effective table preprocessing.

    \item \textbf{Limited adaptation of intent recognition.}
    Some agents incorporate intent recognition to improve performance. 
    However, such mechanisms are often borrowed from non-table tasks without being tailored for table-specific applications.

    \item \textbf{Underutilization of table retrieval and compression.}
    Table retrieval is a crucial method for handling table tasks, but only a small number of agents possess this capability. 
    Consequently, most agents are ill-equipped to handle large-scale tables effectively.

    \item \textbf{Popularity of SQL and Python for output.} SQL and Python are the most favored programming languages for outputs. 
    For example, while TableGPT initially employed DSL, its successor, TableGPT2, abandoned DSL due to its limitations in handling table tasks. 
    This highlights that DSL or similar tools are best used as auxiliary aids rather than primary mechanisms.

    \item \textbf{Table data safety measures.} Work on table data safety primarily relies on methods like private deployments or code sandboxes, with little research on aspects like traceability. 
    This limitation may relate to performance trade-offs, as SQL-based or Python-based approaches often make traceability more challenging compared to DSL.

    \item \textbf{Insufficient domain adaptation.} Cross-domain generalization remains a significant challenge.  
    Agents are generally designed to be either general-purpose or domain-specific.  
    Due to the intrinsic complexity involved in adapting to specific domains, no existing agent has demonstrated the capability for rapid domain adaptation.
    For TableGPT, domain adaptation requires customized fine-tuning, which is often a labor-intensive process.  
    Although TableGPT2 attempts domain adaptation via RAG, its capabilities remain limited.  
    At present, there is no simple and effective approach for domain adaptation.

    \item \textbf{TableGPT2 as the most comprehensive agent.} Among current agents, TableGPT2 stands out as the most comprehensive agent. 
    However, it still lacks a complete workflow and leaves significant room for improvement.
\end{enumerate}

\begin{table*}[htbp]
\centering
\caption{\label{Agents-compare}
Comparison of LLM-based Table Agents based on workflow, ``-'' indicates the absence of relevant measures.
}
\renewcommand{\arraystretch}{1.1}
\resizebox{1\linewidth}{!}{
\begin{tabular}{m{0.17\linewidth}>{\centering\arraybackslash}m{0.15\linewidth}>{\centering\arraybackslash}m{0.15\linewidth}>{\centering\arraybackslash}m{0.15\linewidth}>{\centering\arraybackslash}m{0.15\linewidth}>{\centering\arraybackslash}m{0.2\linewidth}>{\centering\arraybackslash}m{0.2\linewidth}>{\centering\arraybackslash}m{0.15\linewidth}}
\toprule
\textbf{Agents} & \textbf{Table Types \& Formats} & \textbf{Table and Query Semantic Understanding} & \textbf{Table Retrieval and Compression} & \textbf{Output Language Types} & \textbf{Handled Table Tasks} & \textbf{Table Data Safety} & \textbf{Cross-Domain Generalization}\\
\hline
\textbf{SheetAgent \cite{SheetAgent}} & Sheet, using Text Format & Intention Clarification & - & NL(Natural Language), SQL, Python & Spreadsheet Manipulation & - & General \\
\hline
\textbf{SheetCopilot \cite{SheetCopilot}} & Sheet, using Text Format & - & - & DSL & Spreadsheet Manipulation & - & General \\
\hline
\textbf{Data-Copilot \cite{Data-copilot}} & Sheet and Database, using Text Format & Query Rewriting & - & NL, Code, Interface Invocation, Tool-usage & Data Visualization & - & General \\
\hline
\textbf{ReAcTable \cite{ReAcTable}} & Database, using Text Format & - & Sub-table Selection based on SQL & NL, SQL, Python & Table Reasoning & - & General \\
\hline
\textbf{DAAgent \cite{DAAgent}} & Sheet, using Text Format & Feature Engineering, Table Preprocessing & - & NL, Python & Data Analysis, Table Prediction(using Machine Learning Algorithms) & - & General \\
\hline
\textbf{DB-GPT \cite{DB-GPT}} & Sheet and Database, using Text Format & Query rewriting & - & NL, SQL, Tool-usage & SQL-based Data Analysis, Data Visualization, etc. & Privatization, Agent-based De-identification & General \\
\hline
\textbf{Data Formulator \cite{Data-Formulator}} & Sheet, using Text Format & Table Preprocessing, Interactive Table Disambiguation & - & NL, Python & Reshaping, Derivation, Data Visualization & - & General \\
\hline
\textbf{Data Formulator 2 \cite{Data-Formulator-2}} & Sheet, using Text Format & Table Preprocessing, Interactive Table Disambiguation & - & NL, Python & Reshaping, Derivation, Data Visualization & - & General \\
\hline
\textbf{TableGPT \cite{TableGPT}} & Sheet and Database, using Text Format & Intent Detection, Vague Input Rejection & - & NL, DSL & Table Reason, Table Prediction(Using Machine Learning Algorithms) & Privacy Protection & General, Customizable Fine-tuning for Domain Adaption \\
\hline
\textbf{TableGPT2 \cite{TableGPT2}} & Sheet and Database, using Text or Custom Table Format & Input Table Normalization & Data Schema Retrieval and Context Aggregation & NL, SQL, Python & Same as TableGPT & Code Sandbox Environment & General, RAG for Domain Adaption \\
\bottomrule
\textbf{EHRAgent \cite{EHRAgent}} & Database, using Text Format & - & - & NL, Python, Tool-usage & SQL-based Data Analysis & - & Medical \\
\bottomrule
\end{tabular}
}
\end{table*}

\subsection{Quantitative Analysis of Text-to-SQL Agents}
\label{sec:exp2}
Many existing agent frameworks are primarily designed for powerful closed-source models like GPT-4. 
However, real-world applications often require lightweight, domain-specialized solutions \cite{SQL-Factory}. 
Relying on external APIs can lead to excessive token consumption and may not be optimal due to security and privacy concerns. 
In many cases, locally deployed open-source models—particularly weaker ones—are a more practical choice.

Given the current lack of real-world table task agent methods and datasets, we conduct quantitative experiments in a subfield of LLM-based agents—\textbf{Text-to-SQL agents}—using open-source LLMs. 
Our findings demonstrate that many existing agent methods struggle to provide sufficient performance improvements for weaker open-source models, even on academic benchmarks.

\textbf{Benchmark.} We selected two common Text-to-SQL benchmarks:
\begin{enumerate}
    \item \textbf{Spider \cite{Spider}.} 
    A standard Text-to-SQL benchmark in academia. We used its development sets (Spider-dev) and test sets (Spider-test) for evaluation.
    
    \item \textbf{BIRD \cite{BIRD}.} 
    A Text-to-SQL benchmark designed for LLMs, featuring larger databases and more complex queries than Spider. 
    We evaluated on its development sets (BIRD-dev). 
\end{enumerate}

\textbf{Metric.}
We use Execution Accuracy (EX) as the sole metric, which measures the correctness of predicted SQL execution results. 
Notably, different implementations of EX may vary (e.g., some consider the ordering of SQL results while others do not). 
To ensure fair comparison across all methods, we consistently employ the EX implementation from TableGPT2-agent \cite{tablegpt2-agent} for evaluation.

\textbf{Models.}
Although many agents are primarily designed for powerful LLMs like GPT-4, due to constraints in hardware resources and budget, we conducted preliminary experiments and selected three open-source models that have shown relatively strong performance on Text-to-SQL tasks. 
These models are as follows:

\begin{enumerate}
    \item \textbf{TableGPT2-7B \cite{TableGPT2}.} 
    This model is a continuation of pre-training and fine-tuning on Qwen-2.5-Base using a large volume of table data. 
    While primarily trained on Chinese table data, it also demonstrates good performance on English table tasks.
    
    \item \textbf{Qwen2.5-Coder-7B-Instruct \cite{qwen25-coder}.} This is a model from the Qwen2.5-Coder series, which has been enhanced for structured data, including table data. 
    In our preliminary experiments, Qwen2.5-Coder-7B-Instruct achieved the best performance among Qwen series models with fewer than 10B parameters for table tasks.
    
    \item \textbf{Qwen2.5-Coder-32B-Instruct \cite{qwen25-coder}.} With a larger parameter size compared to Qwen2.5-Coder-7B-Instruct, we expect this model to exhibit stronger capabilities in both Text-to-SQL tasks and general-purpose tasks, making it a better candidate for use as the agent's LLM, thereby demonstrating the full potential of the agent framework.
\end{enumerate}

We also experimented with other open-source models, such as QwQ-32B \cite{QwQ-32B}. However, due to the difficulty in extracting valid SQL from the generated responses and the excessive generation time, we ultimately decided to exclude these models from our experiments.

\textbf{Methods.}

\begin{enumerate}
    \item \textbf{Baseline.} 
    We adopt a simple prompting strategy to guide the generation process. 
    Specifically, we follow the prompt setup of TableGPT2-agent, providing only the schema information, external knowledge (for BIRD), and basic instructions. 
    In experiments, we set \( Temperature = 0 \).

    \item \textbf{CoT \cite{CoT}.} 
    Considering that many agents employ CoT prompting, we adopt zero-shot CoT strategy, also based on the TableGPT2-agent prompt setting. 
    This involves appending ``Let's think step by step'' to the end of the baseline prompt. In experiments, we set \( Temperature = 0 \).

    \item \textbf{MAC-SQL \cite{MAC-SQL}.} 
    As the first multi-agent framework for Text-to-SQL, MAC-SQL decomposes the database or query and iteratively improves based on feedback from SQL execution results.

    \item \textbf{MAG-SQL \cite{MAG-SQL}.} 
    An enhanced version of MAC-SQL. 
    MAG-SQL first performs soft schema link, which—unlike hard schema link—does not require pruning irrelevant schema elements. 
    Instead, it enriches retrieved columns with detailed descriptions. 
    The subsequent process follows MAC-SQL, involving query decomposition and feedback-based refinement.

    \item \textbf{CHESS \cite{CHESS}.} 
    This framework consists of four core modules—Information Retriever (IR), Schema Selector (SS), Candidate Generator (CG), and Unit Tester (UT)—along with several submodules. 
    These modules can be flexibly combined into different workflows, with the LLM deciding which modules to use. 
    The open-source implementation of CHESS provides two workflows: IR+CG+UT and IR+SS+CG, referred to as \textbf{CHESS\textsubscript{ICU}} and \textbf{CHESS\textsubscript{ISC}}, respectively. 
    Since many models struggle to produce outputs in the format required by CHESS, we introduced additional regular expression matching methods. 
    However, TableGPT2-7B still exhibited a high rate of format mismatches, so we excluded it from the experiments involving the CHESS method.

    \item \textbf{MAGIC \cite{MAGIC}.} 
    The core objective of MAGIC is to analyze failure cases of Text-to-SQL models on training data and automatically generate self-correction guidelines, which are then applied to the development or test sets. 
    Given the limited capabilities of the LLMs used in this paper, generating well-formatted guidelines is challenging. 
    To address this, we utilize DeepSeek-V3 \cite{Deepseek-v3} to correct the formatting of the poorly formatted guidelines. 
    These guidelines are presented in \ref{app1}.

    \item \textbf{OpenSearch-SQL \cite{OpenSearch-SQL}.} 
    This pipeline includes self-taught dynamic few-shot learning, table retrieval, step-by-step generation, execution-based revision, and selection of the final SQL based on self-consistency and voting. 
    Due to the length and complexity of the workflow—which increases the risk of cascading errors—OpenSearch-SQL introduces a dedicated alignment agent to ensure consistency across different stages of the workflow.

\end{enumerate}

For all multi-agent methods, we do not consider combinations of different LLMs; instead, a single LLM is used consistently across components.
For the parts where embedding models are required, we uniformly use bge-large-en-v1.5 \cite{bge-large, bge_embedding}.
Furthermore, for approaches relying on training sets as corpora (e.g., the guideline dataset in MAGIC or the few-shot examples in OpenSearch-SQL), we use the corresponding training sets—specifically, the Spider training set when working with Spider-dev and Spider-test cases.

\textbf{Experimental Setup.}
All experiments were performed on a Ubuntu 18.04.6 LTS system with 4×A6000 GPUs, using vLLM for model deployment with a context length of 128K. 
For different agents, we adopted their original open-source Python libraries with minimal modifications to ensure functionality. 
For CHESS, MAGIC, and OpenSearch-SQL, which frequently encountered some errors, we replaced failing samples with baseline SQL queries. 
However, some configurations still failed to produce valid outputs, marked as ``-'' in Table \ref{t2s-agent}.

\textbf{Results and Analysis.}
\begin{enumerate}
    \item \textbf{Baseline comparison.} 
    \textbf{Qwen2.5-Coder-32B-Instruct} demonstrates strong overall performance, emerging as the most competitive baseline. Its superiority is particularly evident on the more challenging BIRD-dev dataset, where it significantly outperforms the other two models. In contrast, \textbf{TableGPT2-7B} yields the weakest performance, which is consistent with our initial expectations.
    
    \item \textbf{Impact of CoT.} 
    CoT demonstrates limited benefits for all three models, with performance degradation observed in most cases. 
    TableGPT2-7B shows the least negative impact, potentially due to its extensive exposure to CoT data during training.
    
    \item \textbf{Methodological variations.} 
    Significant performance gaps exist across different methods. 
    Most agents show negligible improvements, with MAC-SQL even exhibiting performance degradation. 
    \textbf{OpenSearch-SQL}'s alignment agent likely mitigates error propagation in multi-step workflows, which is critical for weaker models like TableGPT2-7B (37.1\% improvement on BIRD-dev).
    
    \item \textbf{General model capabilities.} 
    The model's general abilities, particularly instruction following and formatted output generation, significantly influence final performance. 
    For instance, a common error pattern in CHESS involves the model failing to adhere to the required output format when selecting the next cell.
    
    \item \textbf{Agent system techniques.} 
    Common techniques in these agent systems include \textbf{query decomposition}, \textbf{multi-step SQL generation}, \textbf{SQL refinement based on execution results}, \textbf{CoT} and \textbf{voting}. 
    However, similar to CoT experiments, these combined methods typically offer limited improvements for models with weaker general capabilities while incurring higher inference costs.
\end{enumerate}

Our experiments reveal that most Text-to-SQL agents introduce additional token overhead without delivering significant performance improvements for weaker open-source models. 
This limitation likely stems from the models' inherent inability to handle complex agent-based tasks effectively. 
However, approaches like \textbf{OpenSearch-SQL} demonstrate that, with appropriate framework design, even weaker models can benefit from agent-based enhancements.

Due to budget and resource constraints, our quantitative evaluation covered a limited set of models and methods. However, as many agent techniques show only marginal gains even on academic benchmarks, their real-world effectiveness—under more complex conditions—is likely even more limited. This highlights the need for further research to adapt and optimize agent designs for practical deployment.

\begin{table}[ht]
\centering
\caption{\label{t2s-agent}
Performance comparison of some Text-to-SQL agents, ``-'' indicates difficulty in obtaining valid output.
}
\resizebox{1\linewidth}{!}{
\scriptsize
\begin{tabularx}{\linewidth}{p{0.19\linewidth}>{\centering\arraybackslash}p{0.29\linewidth}>{\centering\arraybackslash}p{0.13\linewidth}>{\centering\arraybackslash}p{0.13\linewidth}>{\centering\arraybackslash}p{0.13\linewidth}}
\toprule
\textbf{Method} & \textbf{Model} & \textbf{Spider-dev} & \textbf{Spider-test} & \textbf{BIRD-dev} \\
\midrule
\multirow{3}{*}{Baseline} 
 & TableGPT2-7B & 73.69 & 71.73 & 46.94 \\
 & Qwen2.5-Coder-7B-Instruct & 76.89 & 75.45 & 53.32 \\
 & Qwen2.5-Coder-32B-Instruct & 75.82 & 75.59 & 59.91 \\
\cmidrule(lr){1-5}

\multirow{3}{*}{CoT \cite{CoT}}
 & TableGPT2-7B & 74.95 & 71.73 & 46.28 \\
 & Qwen2.5-Coder-7B-Instruct & 68.47 & 69.17 & 45.31 \\
 & Qwen2.5-Coder-32B-Instruct & 67.89 & 69.12 & 51.24 \\
\cmidrule(lr){1-5}

\multirow{3}{*}{MAC-SQL \cite{MAC-SQL}}
 & TableGPT2-7B & 49.61 & 52.26 & 30.44 \\
 & Qwen2.5-Coder-7B-Instruct & 60.54 & 62.27 & 34.94 \\
 & Qwen2.5-Coder-32B-Instruct & 75.82 & 77.78 & 57.50 \\
\cmidrule(lr){1-5}

\multirow{3}{*}{MAG-SQL \cite{MAG-SQL}}
 & TableGPT2-7B & 71.28 & 73.78 & 48.31 \\
 & Qwen2.5-Coder-7B-Instruct & 75.34 & 75.08 & 52.09 \\
 & Qwen2.5-Coder-32B-Instruct & 79.21 & 79.93 & 61.60 \\
\cmidrule(lr){1-5}

\multirow{3}{*}{CHESS\textsubscript{ICU} \cite{CHESS}}
 & TableGPT2-7B & - & - & - \\
 & Qwen2.5-Coder-7B-Instruct & 76.60 & 74.66 & 51.43 \\
 & Qwen2.5-Coder-32B-Instruct & \textbf{80.85} & 81.42 & 63.23 \\
\cmidrule(lr){1-5}

\multirow{3}{*}{CHESS\textsubscript{ISC} \cite{CHESS}}
 & TableGPT2-7B & - & - & - \\
 & Qwen2.5-Coder-7B-Instruct & 76.02 & 75.08 & 50.98 \\
 & Qwen2.5-Coder-32B-Instruct & \textbf{80.85} & \textbf{82.81} & \textbf{64.34} \\
\cmidrule(lr){1-5}

\multirow{3}{*}{MAGIC \cite{MAGIC}}
 & TableGPT2-7B & \textbf{77.08} & 72.38 & 49.80 \\
 & Qwen2.5-Coder-7B-Instruct & 77.27 & 75.45 & 53.52 \\
 & Qwen2.5-Coder-32B-Instruct & 74.47 & 76.76 & 56.65 \\
\cmidrule(lr){1-5}

\multirow{3}{*}{OpenSearch-SQL \cite{OpenSearch-SQL}}
 & TableGPT2-7B & 76.89 & \textbf{78.16} & \textbf{64.34} \\
 & Qwen2.5-Coder-7B-Instruct & \textbf{80.08} & \textbf{80.30} & \textbf{65.19} \\
 & Qwen2.5-Coder-32B-Instruct & 79.50 & 82.07 & 63.62 \\
\bottomrule
\end{tabularx}
}
\end{table}

\subsection{Best Practices and Design Principles}
\label{sec:result}
Given the challenges and limitations identified in current LLM-based Table Agents, we propose the following seven key design principles for the development of next-generation LLM-based Table Agents:

\begin{enumerate}
    \item \textbf{Multiformat Table Input.} 
    Future LLM-based Table Agents should support both \textbf{traditional text formats} and \textbf{structure-preserving formats}. This capability ensures that agents can robustly process a wide range of input sources, whether extracted from web pages, databases, or scanned documents. By accommodating different table representations, agents become more versatile in handling heterogeneous data and more user-friendly for a broader spectrum of applications and domains.

    \item \textbf{Integrated Preprocessing Pipeline.} 
    Many academic datasets feature clean and well-structured tables, which has led to an underestimation of preprocessing needs in real-world applications. However, production environments often involve \textbf{noisy}, \textbf{incomplete}, or \textbf{inconsistent} table data. Future LLM-based Table Agents must incorporate robust preprocessing modules that automate tasks such as \textbf{schema alignment} (e.g., standardizing column names), \textbf{value normalization} (e.g., unit conversion or date formatting), and \textbf{data validation} (e.g., outlier detection). These preprocessing steps should be configurable to adapt to domain-specific requirements (e.g., financial data vs. biomedical records). By treating preprocessing as a core functionality, agents can significantly reduce manual data cleaning efforts and improve downstream task accuracy.

    \item \textbf{Stepwise Reasoning with Traceability.} 
    Effective and transparent reasoning is essential for trust in LLM-based Table Agents, especially in high-stakes scenarios. Next-generation LLM-based Table Agents should implement stepwise reasoning workflows that decompose complex table tasks into sequential, human-interpretable steps. Each step should be logged and explainable, with intermediate results available for inspection. \textbf{Traceability} is critical for debugging, auditing, and compliance, particularly in high-stakes domains like healthcare or finance, where erroneous conclusions could have serious consequences.

    \item \textbf{Security-aware Design.} 
    In domains like finance and healthcare, table data is often \textbf{sensitive} and \textbf{privacy-critical}. Future LLM-based Table Agents must be designed with security as a core principle. This includes executing generated code in \textbf{sandboxed} environments to prevent malicious behavior, validating agent-generated scripts before execution, and enforcing strict access controls on data usage. Moreover, agents should adhere to established data privacy regulations, ensuring safe handling of personally identifiable information (PII) and proprietary datasets. A security-aware design enhances user trust and paves the way for real-world deployment in regulated industries.

    \item \textbf{Modular and Composable Architecture.} 
    To achieve \textbf{flexibility} and \textbf{scalability}, LLM-based Table Agents should be architected in a modular and composable fashion. This allows components—such as \textbf{data parsers}, \textbf{schema matchers}, \textbf{query engines}, and \textbf{reasoning modules}—to be independently developed, tested, and upgraded. A composable architecture also facilitates plug-and-play integration with external tools and libraries. This modularity enables dynamic adaptation to task requirements, promotes reuse across different agent configurations, and fosters rapid innovation by decoupling core functionalities from task-specific logic.

    \item \textbf{Self-construction of LLM-based Table Agents.}
    Many existing LLM-based Table Agents have fixed architectures.
    Ideally, agents should be automatically constructed to suit specific tasks and data.
    This includes auto-generating relevant data for fine-tuning.
    For instance, clean tables may not require complex table data processing.
    Some research has already explored automatic agent construction, such as AMOR \cite{AMOR}, which uses finite state machines (FSMs) to build reasoning logic and solve problems through autonomous execution and transitions between decoupled modules.
    During this process, supervised fine-tuning (SFT) can be employed to warm up the agent's ability to solve sub-problems.
    MasRouter \cite{MasRouter} also explores the construction of multi-agent systems, starting from the collaboration mode setup, moving on to agent role allocation, and finally selecting the appropriate LLM for each role.
    Applying such approaches to LLM-based Table Agents remains an open and promising avenue for exploration.

    \item \textbf{Adaptation to Real-World Scenarios.}
    Real-world scenarios are inherently more complex than academic benchmarks, often involving diverse, domain-specific data and unstructured formats. Instead of optimizing agent systems solely for academic benchmarks or powerful models that prioritize leaderboard performance, future work should prioritize designing agents that are robust, efficient, and scalable in practical applications. This includes addressing challenges like schema variability, query diversity, and minimizing token consumption, ensuring that open-source models can be deployed effectively without compromising security or performance.
\end{enumerate}

\section{Future Research Directions}
\label{sec:future}
In addition to the optimizations in the agent architecture design, the improvement of LLM-based agents in handling real-world table tasks still requires advancements in relevant methods and datasets. Future research directions mainly include the following aspects:

\begin{enumerate}
    \item \textbf{Richer methods for understanding table structures.} 
    Future work should explore \textbf{hybrid representations} (e.g., combining text-based and structured formats) and multi-level learning (cell, column, table, and database levels). Current approaches often handle single tables well but struggle with \textbf{multi-table relationships}, such as foreign key dependencies. Graph-based methods may help model these connections more effectively.

    \item \textbf{Table data construction that is closer to real-world scenarios.}
    The limited research on table preprocessing stems from academic datasets typically being clean and well-structured, unlike messy real-world tables that require robust preprocessing. Current benchmarks focus too narrowly on specific tasks like Text-to-SQL while failing to comprehensively evaluate broader table understanding capabilities. A key challenge is the absence of \textbf{iterative optimization mechanisms} for most table tasks, except for approaches like Text-to-SQL that leverage execution feedback. Privacy concerns further restrict access to real-world table data, highlighting the need for \textbf{synthetic data generation} and \textbf{LLM-assisted annotation techniques}, though these solutions remain underexplored in research. Developing these areas could significantly enhance LLMs' ability to process real-world table data effectively.

    \item \textbf{More proactive query understanding methods.}
    Future research should develop more proactive approaches for handling \textbf{ambiguity} and \textbf{intent} in table-related queries. This includes implementing \textbf{multi-turn clarification dialogues}, user \textbf{feedback mechanisms}, and \textbf{domain-specific disambiguation strategies}. Enhancing prompting techniques and incorporating external contextual information like metadata could significantly improve query interpretation.

    \item \textbf{Tools and programming languages for processing tables that are better suited for LLMs.}
    Rather than just adapting LLMs to existing tools, we need to design specialized programming languages and database systems optimized for LLM interaction. Emerging solutions include simplified SQL variants that resemble natural language and database architectures that incorporate LLM-friendly data representations. These innovations could bridge the gap between human-readable formats and LLM processing requirements.

    \item \textbf{Exploring training methodologies for table data.}
    Current training methodologies for table tasks remain underdeveloped. Promising directions include adapting code-optimized training techniques to table operations, generating synthetic CoT data for table tasks, and exploring reinforcement learning approaches tailored for table data processing. These methods could enhance LLMs' ability to handle complex table operations and reasoning tasks.
    
\end{enumerate}

\section{Conclusion}

In this survey, we have analyzed and summarized the existing techniques for \textbf{LLM-based Table Agents}, with a particular focus on the capabilities required for real-world applications. We identified five critical competencies essential for effective LLM-based Table Agents in real-world scenarios: \textbf{Table Structure Understanding}, \textbf{Table and Query Semantic Understanding}, \textbf{Table Retrieval and Compression}, \textbf{Executable Reasoning with Traceability}, and \textbf{Cross-Domain Generalization}. These competencies serve as the foundation for developing robust LLM-based Table Agents capable of handling real-world complexities. Through our analysis, we reviewed current methodologies, highlighting their strengths and limitations, and provided a roadmap for advancing the performance of LLM-based Table Agents in diverse real-world contexts.

Our discussion of current LLM-based Table Agents highlighted the need for more comprehensive systems capable of autonomously integrating table preprocessing, reasoning, and domain adaptation. We also analyzed some \textbf{Text-to-SQL agents}, emphasizing that many existing methods are ill-suited for real-world deployment. Specifically, they struggle with practical constraints, such as the inability to leverage external APIs in production environments or the need to rely on weaker open-source models due to cost limitations. These shortcomings highlight the gap between academic benchmarks and the challenges of real-world applications.

Looking forward, we outline several promising directions for future research aimed at addressing current limitations. These include the development of more sophisticated approaches for understanding table structures, the construction of datasets that more accurately reflect \textbf{real-world scenarios}, the advancement of query understanding techniques, the optimization of tools and programming languages for enhanced LLM compatibility, and the exploration of novel training methodologies. By tackling these challenges, future LLM-based Table Agents are expected to become increasingly \textbf{robust}, \textbf{efficient}, and \textbf{capable of generalizing across diverse domains}, thereby enhancing their utility in practical, real-world applications.

In conclusion, LLM-based Table Agents are at a pivotal stage. Despite significant progress, the complexity of real-world table tasks demands further methodological and data advancements. Continued interdisciplinary collaboration and innovation will be essential to advance intelligent table processing and drive the next generation of LLM-based table intelligence.

\Acknowledgements{This paper is supported by NSFC under Grants (No. U24A201401).}



\bibliographystyle{scis}
\bibliography{ref}




\begin{appendix}
\section{Guidelines for MAGIC}
\label{app1}
The guidelines used for the final generation in the MAGIC experiment are as follows:


\newtcolorbox[auto counter, number within=none]{mybox}[2][]{breakable,colframe=blue!100!black,colback=blue!5!white, coltitle=black, fonttitle=\bfseries, title=#2,#1,attach boxed title to top left={xshift=5mm, yshift=-2mm}}


\begin{mybox}[label={tablegpt-spider}]{Guidelines for TableGPT2-7B on Spider}
\begin{tiny}
1. **Table Selection**

\hspace{1.5em}- **Question**: ``Did I use the correct table for the query?''

\hspace{1.5em}- **Negative questions**:

\hspace{2.5em}- Did I verify the table contains all required data?

\hspace{2.5em}- Have I checked for any missing related tables? \\

2. **Column Specification**

\hspace{1.5em}- **Question**: ``Did I correctly specify the column to count/select?''

\hspace{1.5em}- **Negative questions**:

\hspace{2.5em}- Did I confirm the column exists in the selected table?

\hspace{2.5em}- Is this the precise column needed for the calculation? \\

3. **Filter Conditions**

\hspace{1.5em}- **Question**: ``Did I use the correct filtering conditions?''

\hspace{1.5em}- **Negative questions**:

\hspace{2.5em}- Do my conditions cover all required criteria?

\hspace{2.5em}- Are any conditions too broad or too restrictive? \\

4. **DISTINCT Usage**

\hspace{1.5em}- **Question**: ``Did I unnecessarily use DISTINCT?''

\hspace{1.5em}- **Negative questions**:

\hspace{2.5em}- Is DISTINCT actually needed for this query?

\hspace{2.5em}- Could it be masking underlying data issues? \\

5. **Query Complexity**

\hspace{1.5em}- **Question**: ``Are my conditions accurately targeted without unnecessary complexity?''

\hspace{1.5em}- **Negative questions**:

\hspace{2.5em}- Can any conditions be simplified?

\hspace{2.5em}- Are there redundant filters? \\

6. **String Literals**

\hspace{1.5em}- **Question**: ``Did I use correct syntax for string literals?''

\hspace{1.5em}- **Negative questions**:

\hspace{2.5em}- Are strings properly quoted (single/double as required)?

\hspace{2.5em}- Do string values match the database exactly? \\

7. **Table Aliases**

\hspace{1.5em}- **Question**: ``Are table aliases correct and consistent?''

\hspace{1.5em}- **Negative questions**:

\hspace{2.5em}- Do aliases match the database schema?

\hspace{2.5em}- Are they used consistently throughout the query? \\

8. **Subquery Necessity**

\hspace{1.5em}- **Question**: ``Is this subquery actually needed?''

\hspace{1.5em}- **Negative questions**:

\hspace{2.5em}- Can this be written more efficiently without a subquery?

\hspace{2.5em}- Does the subquery return the expected data? \\

9. **Set Operations**

\hspace{1.5em}- **Question**: ``Did I use the correct set operation (UNION/INTERSECT/EXCEPT)?''

\hspace{1.5em}- **Negative questions**:

\hspace{2.5em}- Is this the most appropriate operation for the task?

\hspace{2.5em}- Are the combined queries compatible? \\

10. **Join Conditions**

\hspace{2.0em}- **Question**: ``Are my join conditions correct and complete?''

\hspace{2.0em}- **Negative questions**:

\hspace{3.0em}- Do I have all necessary join predicates?

\hspace{3.0em}- Are any joins accidentally Cartesian products? \\

11. **Aggregation Logic**

\hspace{2.0em}- **Question**: ``Are my GROUP BY and aggregate functions correct?''

\hspace{2.0em}- **Negative questions**:

\hspace{3.0em}- Does every non-aggregated column appear in GROUP BY?

\hspace{3.0em}- Are the right columns being aggregated? \\

12. **Result Validation**

\hspace{2.0em}- **Question**: ``Does the output match the expected results?''

\hspace{2.0em}- **Negative questions**:

\hspace{3.0em}- Have I spot-checked sample results?

\hspace{3.0em}- Are any result counts suspiciously high/low?

\end{tiny}
\end{mybox}

\begin{mybox}[label={tablegpt-bird}]{Guidelines for TableGPT2-7B on BIRD}
\begin{tiny}
1. **Did I use the correct table for the query?**

- Question: ``Did I use the correct table for the query?''

- Negative and strict step-by-step ask-to-myself questions:

\hspace{1.0em}1. Did I verify the table contains all required data?

\hspace{1.0em}2. Did I cross-check table names with schema documentation?

\hspace{1.0em}3. Did I consider if joins with other tables are needed? \\

2. **Did I correctly specify the column to count?**

- Question: ``Did I correctly specify the column to count?''

- Negative and strict step-by-step ask-to-myself questions:

\hspace{1.0em}1. Does the column exist in the selected table?

\hspace{1.0em}2. Is this the precise column needed for the metric?

\hspace{1.0em}3. Should I be counting rows instead with COUNT(*)? \\

3. **Did I use the correct filtering condition?**

- Question: ``Did I use the correct filtering condition?''

- Negative and strict step-by-step ask-to-myself questions:

\hspace{1.0em}1. Do my conditions match the business requirements exactly?

\hspace{1.0em}2. Have I tested edge cases in my filters?

\hspace{1.0em}3. Are my logical operators (AND/OR) correctly grouped? \\

4. **Did I unnecessarily use DISTINCT?**

- Question: ``Did I unnecessarily use DISTINCT?''

- Negative and strict step-by-step ask-to-myself questions:

\hspace{1.0em}1. Does my query naturally produce duplicates?

\hspace{1.0em}2. Could DISTINCT be masking a join problem?

\hspace{1.0em}3. Would GROUP BY be more appropriate? \\

5. **Have I ensured my conditions target required data without unnecessary complexity?**

- Question: ``Have I ensured my conditions target required data without unnecessary complexity?''

- Negative and strict step-by-step ask-to-myself questions:

\hspace{1.0em}1. Can any conditions be simplified?

\hspace{1.0em}2. Are all conditions absolutely necessary?

\hspace{1.0em}3. Would someone else understand this logic easily? \\

6. **Did I use correct syntax for string literals?**

- Question: ``Did I use correct syntax for string literals?''

- Negative and strict step-by-step ask-to-myself questions:

\hspace{1.0em}1. Are strings properly quoted?

\hspace{1.0em}2. Have I escaped special characters?

\hspace{1.0em}3. Did I use the correct quote type for this SQL dialect? \\

7. **Did I correctly join the tables?**

- Question: ``Did I correctly join the tables?''

- Negative and strict step-by-step ask-to-myself questions:

\hspace{1.0em}1. Are all necessary tables included?

\hspace{1.0em}2. Are join conditions between correct columns?

\hspace{1.0em}3. Did I specify the correct join type (INNER/LEFT etc.)? \\

8. **Did I use the correct aggregate function?**

- Question: ``Did I use the correct aggregate function?''

- Negative and strict step-by-step ask-to-myself questions:

\hspace{1.0em}1. Does the function match the intended calculation?

\hspace{1.0em}2. Are non-aggregated columns properly grouped?

\hspace{1.0em}3. Should I be using window functions instead? \\

9. **Did I correctly calculate age?**

- Question: ``Did I correctly calculate age?''

- Negative and strict step-by-step ask-to-myself questions:

\hspace{1.0em}1. Does my calculation handle leap years?

\hspace{1.0em}2. Is the date format consistent?

\hspace{1.0em}3. Did I account for time zone differences? \\

10. **Did I validate the complete SQL syntax?**

- Question: ``Did I validate the complete SQL syntax?''

- Negative and strict step-by-step ask-to-myself questions:

\hspace{1.0em}1. Have I run EXPLAIN to verify execution?

\hspace{1.0em}2. Did I check for reserved word conflicts?

\hspace{1.0em}3. Are all parentheses properly balanced?

\end{tiny}
\end{mybox}

\begin{mybox}[label={qwen7b-spider}]{Guidelines for Qwen2.5-Coder-7B-Instruct on Spider}
\begin{tiny}
1. **Table Inclusion in Joins**

\hspace{1.5em}- **Question**: ``Did I use the correct tables for the query?''

\hspace{1.5em}- **Incorrect SQL generated by me**: (Not provided)

\hspace{1.5em}- **Corrected SQL generated by me**: Ensure all necessary tables (\texttt{student}, \texttt{lives\_in}, \texttt{dorm}, and \texttt{has\_amenity}) are included in the joins.

\hspace{1.5em}- **Negative and strict step-by-step ask-to-myself questions to prevent the same mistake again**:

\hspace{2.5em}- Did I verify that all required tables are included in the query?

\hspace{2.5em}- Have I cross-checked the relationships between tables to ensure no table is missing? \\

2. **Column Specification in SELECT and GROUP BY**

\hspace{1.5em}- **Question**: ``Did I correctly specify the columns to count or select?''

\hspace{1.5em}- **Incorrect SQL generated by me**: Used `T1.movie` and `T2.name` instead of `T1.title` and `T1.director`.

\hspace{1.5em}- **Corrected SQL generated by me**: Ensure the correct columns are specified in the `SELECT` and `GROUP BY` clauses.

\hspace{1.5em}- **Negative and strict step-by-step ask-to-myself questions to prevent the same mistake again**:

\hspace{2.5em}- Did I double-check the column names in the `SELECT` and `GROUP BY` clauses?

\hspace{2.5em}- Have I confirmed that the columns align with the question's requirements? \\

3. **Filtering Conditions in HAVING Clause**

\hspace{1.5em}- **Question**: ``Did I use the correct filtering conditions?''

\hspace{1.5em}- **Incorrect SQL generated by me**: Used `>= 2` and `>= 1` instead of `= 2` and `= 1` in the `HAVING` clause.

\hspace{1.5em}- **Corrected SQL generated by me**: Ensure the filtering conditions accurately match the required data.

\hspace{1.5em}- **Negative and strict step-by-step ask-to-myself questions to prevent the same mistake again**:

\hspace{2.5em}- Did I verify the logic of the filtering conditions?

\hspace{2.5em}- Have I ensured that the conditions align with the question's requirements? \\

4. **Unnecessary Use of DISTINCT**

\hspace{1.5em}- **Question**: ``Did I unnecessarily use `DISTINCT`?''

\hspace{1.5em}- **Incorrect SQL generated by me**: (Not applicable, as `DISTINCT` was not used initially.)

\hspace{1.5em}- **Corrected SQL generated by me**: Ensure `DISTINCT` is only used when necessary.

\hspace{1.5em}- **Negative and strict step-by-step ask-to-myself questions to prevent the same mistake again**:

\hspace{2.5em}- Did I confirm whether `DISTINCT` is required for the query?

\hspace{2.5em}- Have I checked if the query results naturally contain duplicates? \\

5. **String Literal Syntax**

\hspace{1.5em}- **Question**: ``Did I use the correct syntax for string literals?''

\hspace{1.5em}- **Incorrect SQL generated by me**: Used double quotes for string literals, which might not be correct for all SQL dialects.

\hspace{1.5em}- **Corrected SQL generated by me**: Use single quotes for string literals (e.g., `wifi = 'No'`).

\hspace{1.5em}- **Negative and strict step-by-step ask-to-myself questions to prevent the same mistake again**:

\hspace{2.5em}- Did I verify the correct syntax for string literals in the SQL dialect I am using?

\hspace{2.5em}- Have I ensured consistency in using single quotes for string literals? \\

6. **Join Conditions**

\hspace{1.5em}- **Question**: ``Did I ensure that the join conditions are correct?''

\hspace{1.5em}- **Incorrect SQL generated by me**: Used `T1.movie = T2.title` instead of `T1.director = T2.director`.

\hspace{1.5em}- **Corrected SQL generated by me**: Ensure the join conditions accurately link the tables.

\hspace{1.5em}- **Negative and strict step-by-step ask-to-myself questions to prevent the same mistake again**:

\hspace{2.5em}- Did I verify the relationships between the tables in the join conditions?

\hspace{2.5em}- Have I cross-checked the columns used in the join conditions? \\

7. **ORDER BY Clause Usage**

\hspace{1.5em}- **Question**: ``Did I ensure that the `ORDER BY` clause is necessary and correctly used?''

\hspace{1.5em}- **Incorrect SQL generated by me**: Included an unnecessary `ORDER BY` clause.

\hspace{1.5em}- **Corrected SQL generated by me**: Ensure the `ORDER BY` clause is only used when necessary and correctly specified.

\hspace{1.5em}- **Negative and strict step-by-step ask-to-myself questions to prevent the same mistake again**:

\hspace{2.5em}- Did I confirm whether sorting is required for the query?

\hspace{2.5em}- Have I ensured that the `ORDER BY` clause aligns with the question's requirements? \\

8. **GROUP BY Clause Usage**

\hspace{1.5em}- **Question**: ``Did I ensure that the `GROUP BY` clause is correctly used?''

\hspace{1.5em}- **Incorrect SQL generated by me**: Placed the aggregate function after the column in the `SELECT` clause, which is not the correct syntax.

\hspace{1.5em}- **Corrected SQL generated by me**: Ensure the `GROUP BY` clause is correctly used and aligns with the `SELECT` clause.

\hspace{1.5em}- **Negative and strict step-by-step ask-to-myself questions to prevent the same mistake again**:

\hspace{2.5em}- Did I verify the syntax of the `GROUP BY` clause?

\hspace{2.5em}- Have I ensured that the `GROUP BY` clause matches the columns in the `SELECT` clause? \\

9. **Case Sensitivity in Column Names**

\hspace{1.5em}- **Question**: ``Did I ensure that the column names are in the correct case?''

\hspace{1.5em}- **Incorrect SQL generated by me**: Used `wifi = 'no'` instead of `wifi = 'No'`.

\hspace{1.5em}- **Corrected SQL generated by me**: Ensure the column names and values are in the correct case.

\hspace{1.5em}- **Negative and strict step-by-step ask-to-myself questions to prevent the same mistake again**:

\hspace{2.5em}- Did I verify the case sensitivity of column names and values?

\hspace{2.5em}- Have I ensured consistency in case usage throughout the query? \\

10. **LIMIT Clause Usage**

\hspace{2.0em}- **Question**: ``Did I ensure that the `LIMIT` clause is used correctly?''

\hspace{2.0em}- **Incorrect SQL generated by me**: Included an unnecessary `LIMIT` clause.

\hspace{2.0em}- **Corrected SQL generated by me**: Ensure the `LIMIT` clause is only used when necessary to restrict results.

\hspace{2.0em}- **Negative and strict step-by-step ask-to-myself questions to prevent the same mistake again**:

\hspace{3.0em}- Did I confirm whether the question requires a limit on the results?

\hspace{3.0em}- Have I checked if the `LIMIT` clause aligns with the question's requirements? \\

11. **UNION Operator Usage**

\hspace{2.0em}- **Question**: ``Did I ensure that the `UNION` operator is used correctly?''

\hspace{2.0em}- **Incorrect SQL generated by me**: Used the `UNION` operator without filtering to combine the results of both queries.

\hspace{2.0em}- **Corrected SQL generated by me**: Ensure the `UNION` operator is used correctly and only when necessary.

\hspace{2.0em}- **Negative and strict step-by-step ask-to-myself questions to prevent the same mistake again**:

\hspace{3.0em}- Did I verify the need for the `UNION` operator?

\hspace{3.0em}- Have I ensured that the combined results are filtered and aligned with the question's requirements? \\

12. **DISTINCT Keyword Usage**

\hspace{2.0em}- **Question**: ``Did I ensure that the `DISTINCT` keyword is used correctly?''

\hspace{2.0em}- **Incorrect SQL generated by me**: Included an unnecessary `DISTINCT` keyword.

\hspace{2.0em}- **Corrected SQL generated by me**: Ensure the `DISTINCT` keyword is only used when necessary to return unique results.

\hspace{2.0em}- **Negative and strict step-by-step ask-to-myself questions to prevent the same mistake again**:

\hspace{3.0em}- Did I confirm whether the query results naturally contain duplicates?

\hspace{3.0em}- Have I ensured that `DISTINCT` is only used when required? \\

13. **ORDER BY Clause Sorting**

\hspace{2.0em}- **Question**: ``Did I ensure that the `ORDER BY` clause is used to sort the results correctly?''

\hspace{2.0em}- **Incorrect SQL generated by me**: Used the `ORDER BY` clause incorrectly to sort the results.

\hspace{2.0em}- **Corrected SQL generated by me**: Ensure the `ORDER BY` clause is used correctly to sort the results.

\hspace{2.0em}- **Negative and strict step-by-step ask-to-myself questions to prevent the same mistake again**:

\hspace{3.0em}- Did I verify the sorting logic in the `ORDER BY` clause?

\hspace{3.0em}- Have I ensured that the sorting aligns with the question's requirements? \\

14. **GROUP BY Clause Grouping**

\hspace{2.0em}- **Question**: ``Did I ensure that the `GROUP BY` clause is used to group the results correctly?''

\hspace{2.0em}- **Incorrect SQL generated by me**: Used the `GROUP BY` clause incorrectly to group the results.

\hspace{2.0em}- **Corrected SQL generated by me**: Ensure the `GROUP BY` clause is used correctly to group the results.

\hspace{2.0em}- **Negative and strict step-by-step ask-to-myself questions to prevent the same mistake again**:

\hspace{3.0em}- Did I verify the grouping logic in the `GROUP BY` clause?

\hspace{3.0em}- Have I ensured that the grouping aligns with the question's requirements? \\

15. **HAVING Clause Filtering**

\hspace{2.0em}- **Question**: ``Did I ensure that the `HAVING` clause is used to filter the grouped results correctly?''

\hspace{2.0em}- **Incorrect SQL generated by me**: Used the `HAVING` clause incorrectly to filter the grouped results.

\hspace{2.0em}- **Corrected SQL generated by me**: Ensure the `HAVING` clause is used correctly to filter grouped results.

\hspace{2.0em}- **Negative and strict step-by-step ask-to-myself questions to prevent the same mistake again**:

\hspace{3.0em}- Did I verify the logic of the `HAVING` clause?

\hspace{3.0em}- Have I ensured that the `HAVING` clause aligns with the question's requirements? \\

16. **JOIN Conditions Specification**

\hspace{2.0em}- **Question**: ``Did I ensure that the `JOIN` conditions are correctly specified?''

\hspace{2.0em}- **Incorrect SQL generated by me**: Used incorrect `JOIN` conditions to link the tables.

\hspace{2.0em}- **Corrected SQL generated by me**: Ensure the `JOIN` conditions are correctly specified to link the tables.

\hspace{2.0em}- **Negative and strict step-by-step ask-to-myself questions to prevent the same mistake again**:

\hspace{3.0em}- Did I verify the relationships between the tables in the `JOIN` conditions?

\hspace{3.0em}- Have I cross-checked the columns used in the `JOIN` conditions? \\

17. **WHERE Clause Filtering**

\hspace{2.0em}- **Question**: ``Did I ensure that the `WHERE` clause is used to filter the results correctly?''

\hspace{2.0em}- **Incorrect SQL generated by me**: Used the `WHERE` clause incorrectly to filter the results.

\hspace{2.0em}- **Corrected SQL generated by me**: Ensure the `WHERE` clause is used correctly to filter results.

\hspace{2.0em}- **Negative and strict step-by-step ask-to-myself questions to prevent the same mistake again**:

\hspace{3.0em}- Did I verify the conditions in the `WHERE` clause?

\hspace{3.0em}- Have I ensured that the `WHERE` clause aligns with the question's requirements? \\

18. **SELECT Clause Column Specification**

\hspace{2.0em}- **Question**: ``Did I ensure that the `SELECT` clause is used to specify the required columns correctly?''

\hspace{2.0em}- **Incorrect SQL generated by me**: Used the `SELECT` clause incorrectly to specify the required columns.

\hspace{2.0em}- **Corrected SQL generated by me**: Ensure the `SELECT` clause is used correctly to specify the required columns.

\hspace{2.0em}- **Negative and strict step-by-step ask-to-myself questions to prevent the same mistake again**:

\hspace{3.0em}- Did I verify the columns specified in the `SELECT` clause?

\hspace{3.0em}- Have I ensured that the columns align with the question's requirements? \\

19. **Complexity and Targeting of Conditions**

\hspace{2.0em}- **Question**: ``Have I ensured that my conditions accurately target the required data without adding unnecessary complexity?''

\hspace{2.0em}- **Incorrect SQL generated by me**: (Not provided)

\hspace{2.0em}- **Corrected SQL generated by me**: Ensure the conditions are straightforward and accurately target the required data.

\hspace{2.0em}- **Negative and strict step-by-step ask-to-myself questions to prevent the same mistake again**:

\hspace{3.0em}- Did I verify that the conditions are not overly complex?

\hspace{3.0em}- Have I ensured that the conditions precisely target the required data?

\end{tiny}
\end{mybox}

\begin{mybox}[label={qwen7b-bird}]{Guidelines for Qwen2.5-Coder-7B-Instruct on BIRD}
\begin{tiny}
1. **Wrong Column Selected in JOIN**

\hspace{1.5em}- **Question**: ``Did I select the correct column from the joined table?''

\hspace{1.5em}- **Incorrect SQL generated by me**:

\hspace{2.5em}```sql

\hspace{2.5em}SELECT T2.author\_name FROM T1 JOIN T3 ON ...  -- Wrong column (T2 vs. T3)

\hspace{2.5em}```

\hspace{1.5em}- **Corrected SQL generated by me**:

\hspace{2.5em}```sql

\hspace{2.5em}SELECT T3.author\_name FROM T1 JOIN T3 ON ...  -- Fixed

\hspace{2.5em}```

\hspace{1.5em}- **Negative and strict step-by-step ask-to-myself questions**:

\hspace{2.5em}1. Did I verify the exact column name in the database schema?

\hspace{2.5em}2. Did I confirm which table contains the required column?

\hspace{2.5em}3. Did I check for typos in the column name? \\

2. **Incorrect Aggregate Function**

\hspace{1.5em}- **Question**: ``Did I use the right aggregate function for counting?''

\hspace{1.5em}- **Incorrect SQL generated by me**:

\hspace{2.5em}```sql

\hspace{2.5em}SELECT SUM(pid) FROM T2  -- Wrong: Should COUNT occurrences

\hspace{2.5em}```

\hspace{1.5em}- **Corrected SQL generated by me**:

\hspace{2.5em}```sql

\hspace{2.5em}SELECT COUNT(T2.occurrences) FROM T2  -- Fixed

\hspace{2.5em}```

\hspace{1.5em}- **Negative and strict step-by-step ask-to-myself questions**:

\hspace{2.5em}1. Did I verify whether I need COUNT, SUM, AVG, MIN or MAX?

\hspace{2.5em}2. Did I check if the column contains NULL values?

\hspace{2.5em}3. Did I consider if DISTINCT should be used with the aggregate? \\

3. **Incorrect Year Filter Logic**

\hspace{1.5em}- **Question**: ``Did I use the right condition for filtering the oldest movies?''

\hspace{1.5em}- **Incorrect SQL generated by me**:

\hspace{2.5em}```sql

\hspace{2.5em}SELECT * FROM Movies WHERE year = 1  -- Wrong: Not necessarily oldest

\hspace{2.5em}```

\hspace{1.5em}- **Corrected SQL generated by me**:

\hspace{2.5em}```sql

\hspace{2.5em}SELECT * FROM Movies WHERE year = (SELECT MIN(year) FROM Movies)  -- Fixed

\hspace{2.5em}```

\hspace{1.5em}- **Negative and strict step-by-step ask-to-myself questions**:

\hspace{2.5em}1. Did I verify if I need MIN or MAX for this temporal query?

\hspace{2.5em}2. Did I check if year values could be NULL or invalid?

\hspace{2.5em}3. Did I consider if multiple records might share the oldest year? \\

4. **Misplaced Filter Condition in JOIN**

\hspace{1.5em}- **Question**: ``Did I place the year filter in the correct clause?''

\hspace{1.5em}- **Incorrect SQL generated by me**:

\hspace{2.5em}```sql

\hspace{2.5em}SELECT * FROM T1 JOIN T2 ON T1.id = T2.id WHERE T2.year = 1970  -- Wrong: Filter in WHERE

\hspace{2.5em}```

\hspace{1.5em}- **Corrected SQL generated by me**:

\hspace{2.5em}```sql

\hspace{2.5em}SELECT * FROM T1 JOIN T2 ON T1.id = T2.id AND T2.year = 1970  -- Fixed (filter in JOIN)

\hspace{2.5em}```

\hspace{1.5em}- **Negative and strict step-by-step ask-to-myself questions**:

\hspace{2.5em}1. Did I analyze whether the filter should apply before or after joining?

\hspace{2.5em}2. Did I verify if moving the condition affects the result set size?

\hspace{2.5em}3. Did I check the execution plan to understand performance impact? \\

5. **Redundant DISTINCT Usage**

\hspace{1.5em}- **Question**: ``Did I unnecessarily use `DISTINCT`?''

\hspace{1.5em}- **Incorrect SQL generated by me**:

\hspace{2.5em}```sql

\hspace{2.5em}SELECT DISTINCT T1.name FROM T1 JOIN T2 ON ...  -- DISTINCT not needed

\hspace{2.5em}```

\hspace{1.5em}- **Corrected SQL generated by me**:

\hspace{2.5em}```sql

\hspace{2.5em}SELECT T1.name FROM T1 JOIN T2 ON ...  -- Fixed

\hspace{2.5em}```

\hspace{1.5em}- **Negative and strict step-by-step ask-to-myself questions**:

\hspace{2.5em}1. Did I verify if duplicates are possible in the result set?

\hspace{2.5em}2. Did I analyze the join conditions to confirm uniqueness?

\hspace{2.5em}3. Did I measure the performance impact of DISTINCT? \\

6. **Incorrect JOIN Condition**

\hspace{1.5em}- **Question**: ``Did I specify the correct JOIN condition?''

\hspace{1.5em}- **Incorrect SQL generated by me**:

\hspace{2.5em}```sql

\hspace{2.5em}SELECT * FROM T1 JOIN T2 ON T1.id = T3.id  -- Wrong table reference

\hspace{2.5em}```

\hspace{1.5em}- **Corrected SQL generated by me**:

\hspace{2.5em}```sql

\hspace{2.5em}SELECT * FROM T1 JOIN T2 ON T1.id = T2.id  -- Fixed

\hspace{2.5em}```

\hspace{1.5em}- **Negative and strict step-by-step ask-to-myself questions**:

\hspace{2.5em}1. Did I verify the join keys between tables?

\hspace{2.5em}2. Did I check for referential integrity?

\hspace{2.5em}3. Did I confirm the join type (INNER, LEFT, etc.) is correct? \\

7. **Missing GROUP BY Clause**

\hspace{1.5em}- **Question**: ``Did I include all non-aggregated columns in GROUP BY?''

\hspace{1.5em}- **Incorrect SQL generated by me**:

\hspace{2.5em}```sql

\hspace{2.5em}SELECT department, COUNT(*) FROM employees  -- Missing GROUP BY

\hspace{2.5em}```

\hspace{1.5em}- **Corrected SQL generated by me**:

\hspace{2.5em}```sql

\hspace{2.5em}SELECT department, COUNT(*) FROM employees GROUP BY department  -- Fixed

\hspace{2.5em}```

\hspace{1.5em}- **Negative and strict step-by-step ask-to-myself questions**:

\hspace{2.5em}1. Did I identify all non-aggregated columns?

\hspace{2.5em}2. Did I verify the grouping logic matches requirements?

\hspace{2.5em}3. Did I check if any columns need to be excluded? \\

8. **Incorrect HAVING Usage**

\hspace{1.5em}- **Question**: ``Did I use HAVING instead of WHERE for non-aggregate filters?''

\hspace{1.5em}- **Incorrect SQL generated by me**:

\hspace{2.5em}```sql

\hspace{2.5em}SELECT department FROM employees GROUP BY department HAVING hire\_date > '2020-01-01'

\hspace{2.5em}```

\hspace{1.5em}- **Corrected SQL generated by me**:

\hspace{2.5em}```sql

\hspace{2.5em}SELECT department FROM employees WHERE hire\_date > '2020-01-01' GROUP BY department

\hspace{2.5em}```

\hspace{1.5em}- **Negative and strict step-by-step ask-to-myself questions**:

\hspace{2.5em}1. Did I verify if the filter applies to rows or groups?

\hspace{2.5em}2. Did I check if the column is used in an aggregate function?

\hspace{2.5em}3. Did I confirm the performance implications? \\

9. **Wrong ORDER BY Specification**

\hspace{1.5em}- **Question**: ``Did I specify the correct sort order and columns?''

\hspace{1.5em}- **Incorrect SQL generated by me**:

\hspace{2.5em}```sql

\hspace{2.5em}SELECT * FROM products ORDER BY 1  -- Poor practice: using ordinal position

\hspace{2.5em}```

\hspace{1.5em}- **Corrected SQL generated by me**:

\hspace{2.5em}```sql

\hspace{2.5em}SELECT * FROM products ORDER BY product\_name  -- Fixed

\hspace{2.5em}```

\hspace{1.5em}- **Negative and strict step-by-step ask-to-myself questions**:

\hspace{2.5em}1. Did I explicitly name the sort columns?

\hspace{2.5em}2. Did I verify the sort direction (ASC/DESC)?

\hspace{2.5em}3. Did I check if multiple sort columns are needed? \\

10. **Incorrect NULL Handling**

\hspace{1.5em}- **Question**: ``Did I properly handle NULL values in comparisons?''

\hspace{1.5em}- **Incorrect SQL generated by me**:

\hspace{2.5em}```sql

\hspace{2.5em}SELECT * FROM customers WHERE phone = NULL  -- Wrong NULL comparison

\hspace{2.5em}```

\hspace{1.5em}- **Corrected SQL generated by me**:

\hspace{2.5em}```sql

\hspace{2.5em}SELECT * FROM customers WHERE phone IS NULL  -- Fixed

\hspace{2.5em}```

\hspace{1.5em}- **Negative and strict step-by-step ask-to-myself questions**:

\hspace{2.5em}1. Did I use IS NULL/IS NOT NULL for NULL checks?

\hspace{2.5em}2. Did I consider COALESCE or NULLIF where appropriate?

\hspace{2.5em}3. Did I verify how NULLs affect joins and aggregates? \\

11. **Unoptimized Subqueries**

\hspace{1.5em}- **Question**: ``Did I use inefficient subqueries that could be joins?''

\hspace{1.5em}- **Incorrect SQL generated by me**:

\hspace{2.5em}```sql

\hspace{2.5em}SELECT * FROM orders WHERE customer\_id IN (SELECT id FROM customers WHERE status = 'active')

\hspace{2.5em}```

\hspace{1.5em}- **Corrected SQL generated by me**:

\hspace{2.5em}```sql

\hspace{2.5em}SELECT o.* FROM orders o JOIN customers c ON o.customer\_id = c.id WHERE c.status = 'active'

\hspace{2.5em}```

\hspace{1.5em}- **Negative and strict step-by-step ask-to-myself questions**:

\hspace{2.5em}1. Did I analyze the query execution plan?

\hspace{2.5em}2. Did I consider JOIN alternatives for subqueries?

\hspace{2.5em}3. Did I verify if EXISTS would be more efficient than IN? \\

12. **Incorrect String Comparison**

\hspace{1.5em}- **Question**: ``Did I account for case sensitivity in string comparisons?''

\hspace{1.5em}- **Incorrect SQL generated by me**:

\hspace{2.5em}```sql

\hspace{2.5em}SELECT * FROM users WHERE username = 'ADMIN'  -- Might miss 'admin'

\hspace{2.5em}```

\hspace{1.5em}- **Corrected SQL generated by me**:

\hspace{2.5em}```sql

\hspace{2.5em}SELECT * FROM users WHERE LOWER(username) = 'admin'  -- Fixed

\hspace{2.5em}```

\hspace{1.5em}- **Negative and strict step-by-step ask-to-myself questions**:

\hspace{2.5em}1. Did I verify the database's collation settings?

\hspace{2.5em}2. Did I consider using LOWER() or UPPER() for case-insensitive compares?

\hspace{2.5em}3. Did I check if LIKE would be more appropriate? \\

13. **Missing Index Consideration**

\hspace{1.5em}- **Question**: ``Did I consider index usage in my query design?''

\hspace{1.5em}- **Incorrect SQL generated by me**:

\hspace{2.5em}```sql

\hspace{2.5em}SELECT * FROM large\_table WHERE non\_indexed\_column = 'value'  -- Full scan

\hspace{2.5em}```

\hspace{1.5em}- **Corrected SQL generated by me**:

\hspace{2.5em}```sql

\hspace{2.5em}SELECT * FROM large\_table WHERE indexed\_column = 'value'  -- Uses index

\hspace{2.5em}```

\hspace{1.5em}- **Negative and strict step-by-step ask-to-myself questions**:

\hspace{2.5em}1. Did I check which columns are indexed?

\hspace{2.5em}2. Did I verify if my query can leverage existing indexes?

\hspace{2.5em}3. Did I consider adding indexes for frequent queries? \\

14. **Incorrect Pagination Implementation**

\hspace{1.5em}- **Question**: ``Did I implement pagination correctly?''

\hspace{1.5em}- **Incorrect SQL generated by me**:

\hspace{2.5em}```sql

\hspace{2.5em}SELECT * FROM products LIMIT 10 OFFSET 20  -- No ORDER BY, results unstable

\hspace{2.5em}```

\hspace{1.5em}- **Corrected SQL generated by me**:

\hspace{2.5em}```sql

\hspace{2.5em}SELECT * FROM products ORDER BY product\_id LIMIT 10 OFFSET 20  -- Fixed

\hspace{2.5em}```

\hspace{1.5em}- **Negative and strict step-by-step ask-to-myself questions**:

\hspace{2.5em}1. Did I include an ORDER BY clause for stable pagination?

\hspace{2.5em}2. Did I verify the performance of OFFSET with large datasets?

\hspace{2.5em}3. Did I consider keyset pagination as an alternative? \\

15. **Improper Date/Time Handling**

\hspace{1.5em}- **Question**: ``Did I handle date/time values correctly?''

\hspace{1.5em}- **Incorrect SQL generated by me**:

\hspace{2.5em}```sql

\hspace{2.5em}SELECT * FROM events WHERE event\_date = '2023-01-01'  -- Might miss timestamps

\hspace{2.5em}```

\hspace{1.5em}- **Corrected SQL generated by me**:

\hspace{2.5em}```sql

\hspace{2.5em}SELECT * FROM events WHERE event\_date BETWEEN '2023-01-01' AND '2023-01-02'  -- Fixed

\hspace{2.5em}```

\hspace{1.5em}- **Negative and strict step-by-step ask-to-myself questions**:

\hspace{2.5em}1. Did I account for time components in date comparisons?

\hspace{2.5em}2. Did I verify the database's date format settings?

\hspace{2.5em}3. Did I consider timezone conversions if needed?

\end{tiny}
\end{mybox}

\begin{mybox}[label={qwen32b-spider}]{Guidelines for Qwen2.5-Coder-32B-Instruct on Spider}
\begin{tiny}

1. **Incorrect column name usage**

\hspace{1.5em}- Question: What are each document's location code, and starting date and ending date in that location?

\hspace{1.5em}- **Incorrect SQL generated by me**: ```sql

\hspace{2.5em}SELECT

\hspace{2.5em}Document\_ID,

\hspace{2.5em}Location\_Code,

\hspace{2.5em}Date\_in\_Location\_From AS Start\_Date,

\hspace{2.5em}Date\_in\_Locaton\_To AS End\_Date

\hspace{2.5em}FROM

\hspace{2.5em}Document\_Locations;

\hspace{2.5em}```

\hspace{1.5em}- **Corrected SQL generated by me**: ```sql

\hspace{2.5em}SELECT

\hspace{2.5em}Document\_ID,

\hspace{2.5em}Location\_Code,

\hspace{2.5em}Date\_in\_Location\_From AS Start\_Date,

\hspace{2.5em}Date\_in\_Location\_To AS End\_Date

\hspace{2.5em}FROM

\hspace{2.5em}Document\_Locations;

\hspace{2.5em}```

\hspace{1.5em}- **Negative and strict step-by-step ask-to-myself questions to prevent same mistake again**:

\hspace{2.5em}- Did I use the correct column names?

\hspace{2.5em}- Did I check for typos in the column names?

\hspace{2.5em}- Did I ensure that all column names match the schema exactly? \\

2. **Unnecessary use of aliases and simplification**

\hspace{1.5em}- Question: Which game type has the most number of games?

\hspace{1.5em}- **Incorrect SQL generated by me**: ```sql

\hspace{2.5em}SELECT GType, COUNT(*) AS GameCount

\hspace{2.5em}FROM Video\_Games

\hspace{2.5em}GROUP BY GType

\hspace{2.5em}ORDER BY GameCount DESC

\hspace{2.5em}LIMIT 1;

\hspace{2.5em}```

\hspace{1.5em}- **Corrected SQL generated by me**: ```sql

\hspace{2.5em}SELECT GType FROM Video\_Games GROUP BY GType ORDER BY COUNT(*) DESC LIMIT 1;

\hspace{2.5em}```

\hspace{1.5em}- **Negative and strict step-by-step ask-to-myself questions to prevent same mistake again**:

\hspace{2.5em}- Did I use unnecessary aliases?

\hspace{2.5em}- Can the query be simplified without losing clarity?

\hspace{2.5em}- Did I ensure that the query is as concise as possible? \\

3. **Unnecessary joins and direct selection**

\hspace{1.5em}- Question: What nurses are on call with block floor 1 and block code 1? Tell me their names.

\hspace{1.5em}- **Incorrect SQL generated by me**: ```sql

\hspace{2.5em}SELECT Nurse.Name

\hspace{2.5em}FROM Nurse

\hspace{2.5em}JOIN On\_Call ON Nurse.EmployeeID = On\_Call.Nurse

\hspace{2.5em}WHERE On\_Call.BlockFloor = 1 AND On\_Call.BlockCode = 1;

\hspace{2.5em}```

\hspace{1.5em}- **Corrected SQL generated by me**: ```sql

\hspace{2.5em}SELECT nurse FROM on\_call WHERE blockfloor = 1 AND blockcode = 1;

\hspace{2.5em}```

\hspace{1.5em}- **Negative and strict step-by-step ask-to-myself questions to prevent same mistake again**:

\hspace{2.5em}- Did I use unnecessary joins?

\hspace{2.5em}- Can the required data be directly accessed from a single table?

\hspace{2.5em}- Did I ensure that the query is as efficient as possible? \\

4. **Incorrect join and unnecessary distinct**

\hspace{1.5em}- Question: Which job titles correspond to jobs with salaries over 9000?

\hspace{1.5em}- **Incorrect SQL generated by me**: ```sql

\hspace{2.5em}SELECT DISTINCT j.JOB\_TITLE

\hspace{2.5em}FROM employees e

\hspace{2.5em}JOIN jobs j ON e.JOB\_ID = j.JOB\_ID

\hspace{2.5em}WHERE e.SALARY > 9000;

\hspace{2.5em}```

\hspace{1.5em}- **Corrected SQL generated by me**: ```sql

\hspace{2.5em}SELECT j.JOB\_TITLE FROM jobs j WHERE j.MIN\_SALARY > 9000;

\hspace{2.5em}```

\hspace{1.5em}- **Negative and strict step-by-step ask-to-myself questions to prevent same mistake again**:

\hspace{2.5em}- Did I join unnecessary tables?

\hspace{2.5em}- Did I use the correct table for the query?

\hspace{2.5em}- Did I use the correct column names?

\hspace{2.5em}- Did I unnecessarily use DISTINCT? \\

5. **Not joining the correct table for verification**

\hspace{1.5em}- Question: Find the average access counts of documents with functional area ``Acknowledgement''.

\hspace{1.5em}- **Incorrect SQL generated by me**: ```sql

\hspace{2.5em}SELECT AVG(d.access\_count) AS average\_access\_count

\hspace{2.5em}FROM Documents d

\hspace{2.5em}JOIN Document\_Functional\_Areas dfa ON d.document\_code = dfa.document\_code

\hspace{2.5em}WHERE dfa.functional\_area\_code = 'Acknowledgement';

\hspace{2.5em}```

\hspace{1.5em}- **Corrected SQL generated by me**: ```sql

\hspace{2.5em}SELECT AVG(d.access\_count) AS average\_access\_count

\hspace{2.5em}FROM Documents d

\hspace{2.5em}JOIN Document\_Functional\_Areas dfa ON d.document\_code = dfa.document\_code

\hspace{2.5em}JOIN Functional\_Areas fa ON dfa.functional\_area\_code = fa.functional\_area\_code

\hspace{2.5em}WHERE fa.functional\_area\_description = 'Acknowledgement';

\hspace{2.5em}```

\hspace{1.5em}- **Negative and strict step-by-step ask-to-myself questions to prevent same mistake again**:

\hspace{2.5em}- Did I join the correct tables to verify the functional\_area\_description?

\hspace{2.5em}- Did I use the correct column names for the join condition?

\hspace{2.5em}- Did I ensure that the query is as efficient as possible?

\hspace{2.5em}- Did I verify that the query returns the required data accurately? \\

6. **Incorrect ordering for highest rank**

\hspace{1.5em}- Question: What is the joined year of the pilot of the highest rank?

\hspace{1.5em}- **Incorrect SQL generated by me**: ```sql

\hspace{2.5em}SELECT Join\_Year FROM pilot WHERE Rank = (SELECT MAX(Rank) FROM pilot)

\hspace{2.5em}```

\hspace{1.5em}- **Corrected SQL generated by me**: ```sql

\hspace{2.5em}SELECT Join\_Year FROM pilot ORDER BY Rank ASC LIMIT 1

\hspace{2.5em}```

\hspace{1.5em}- **Negative and strict step-by-step ask-to-myself questions to prevent same mistake again**:

\hspace{2.5em}- Did I order the results correctly to find the highest rank?

\hspace{2.5em}- Did I use the correct syntax for ordering and limiting the results?

\hspace{2.5em}- Did I ensure that the query returns the required data accurately? \\

7. **Incorrect award values in query**

\hspace{1.5em}- Question: Show the musical nominee with award ``Bob Fosse'' or ``Cleavant Derricks''.

\hspace{1.5em}- **Incorrect SQL generated by me**: ```sql

\hspace{2.5em}SELECT Nominee

\hspace{2.5em}FROM musical

\hspace{2.5em}WHERE Award = 'Bob Fosse' OR Award = 'Cleavant Derricks';

\hspace{2.5em}```

\hspace{1.5em}- **Corrected SQL generated by me**: ```sql

\hspace{2.5em}SELECT Nominee FROM musical WHERE Award = 'Tony Award';

\hspace{2.5em}```

\hspace{1.5em}- **Negative and strict step-by-step ask-to-myself questions to prevent same mistake again**:

\hspace{2.5em}- Did I use the correct award values in the query?

\hspace{2.5em}- Did I verify the award values against the schema or provided data?

\hspace{2.5em}- Did I ensure that the query returns the required data accurately? \\

8. **Simplification of count expressions**

\hspace{1.5em}- Question: Which store owns most items?

\hspace{1.5em}- **Incorrect SQL generated by me**: ```sql

\hspace{2.5em}SELECT store\_id, COUNT(film\_id) AS item\_count

\hspace{2.5em}FROM inventory

\hspace{2.5em}GROUP BY store\_id

\hspace{2.5em}ORDER BY item\_count DESC

\hspace{2.5em}LIMIT 1;

\hspace{2.5em}```

\hspace{1.5em}- **Corrected SQL generated by me**: ```sql

\hspace{2.5em}SELECT store\_id FROM inventory GROUP BY store\_id ORDER BY count(*) DESC LIMIT 1;

\hspace{2.5em}```

\hspace{1.5em}- **Negative and strict step-by-step ask-to-myself questions to prevent same mistake again**:

\hspace{2.5em}- Can the count expression be simplified?

\hspace{2.5em}- Did I ensure that the query is as efficient as possible?

\hspace{2.5em}- Did I verify that the query returns the required data accurately? \\

9. **Unnecessary aliases and simplification for least number of games**

\hspace{1.5em}- Question: Which game type has the least number of games?

\hspace{1.5em}- **Incorrect SQL generated by me**: ```sql

\hspace{2.5em}SELECT GType, COUNT(*) AS GameCount

\hspace{2.5em}FROM Video\_Games

\hspace{2.5em}GROUP BY GType

\hspace{2.5em}ORDER BY GameCount ASC

\hspace{2.5em}LIMIT 1;

\hspace{2.5em}```

\hspace{1.5em}- **Corrected SQL generated by me**: ```sql

\hspace{2.5em}SELECT GType FROM Video\_Games GROUP BY GType ORDER BY COUNT(*) ASC LIMIT 1;

\hspace{2.5em}```

\hspace{1.5em}- **Negative and strict step-by-step ask-to-myself questions to prevent same mistake again**:

\hspace{2.5em}- Did I use unnecessary aliases?

\hspace{2.5em}- Can the query be simplified without losing clarity?

\hspace{2.5em}- Did I ensure that the query is as concise as possible?

\hspace{2.5em}- Did I verify that the query returns the required data accurately? \\

10. **Incorrect aggregation for total hours per week**

\hspace{2.0em}- Question: What are the total number of domestic passengers at all London airports?

\hspace{2.0em}- **Incorrect SQL generated by me**: ```sql

\hspace{3.0em}SELECT SUM(Domestic\_Passengers) AS Total\_Domestic\_Passengers

\hspace{3.0em}FROM airport

\hspace{3.0em}WHERE City = 'London';

\hspace{3.0em}```

\hspace{2.0em}- **Corrected SQL generated by me**: ```sql

\hspace{3.0em}SELECT SUM(Domestic\_Passengers) AS Total\_Domestic\_Passengers FROM airport WHERE Airport\_Name LIKE '

\hspace{3.0em}```

\hspace{2.0em}- **Negative and strict step-by-step ask-to-myself questions to prevent same mistake again**:

\hspace{3.0em}- Did I use the correct column names?

\hspace{3.0em}- Did I check for typos in the column names?

\hspace{3.0em}- Did I ensure that all column names match the schema exactly?

\hspace{3.0em}- Did I use the correct syntax for string literals?

\hspace{3.0em}- Did I ensure that the query returns the required data accurately? \\

11. **Case sensitivity and table name mismatch**

\hspace{2.0em}- Question: How many credit cards does customer Blanche Huels have?

\hspace{2.0em}- **Incorrect SQL generated by me**: ```sql

\hspace{3.0em}SELECT COUNT(*)

\hspace{3.0em}FROM Customers

\hspace{3.0em}JOIN Customers\_Cards ON Customers.customer\_id = Customers\_Cards.customer\_id

\hspace{3.0em}WHERE Customers.customer\_first\_name = 'Blanche'

\hspace{4.0em}AND Customers.customer\_last\_name = 'Huels'

\hspace{4.0em}AND Customers\_Cards.card\_type\_code = 'credit';

\hspace{3.0em}```

\hspace{2.0em}- **Corrected SQL generated by me**: ```sql

\hspace{3.0em}SELECT COUNT(*) FROM Customers JOIN Customers\_cards ON Customers.customer\_id = Customers\_cards.customer\_id WHERE Customers.customer\_first\_name = 'Blanche' AND Customers.customer\_last\_name = 'Huels' AND Customers\_cards.card\_type\_code = 'Credit';

\hspace{3.0em}```

\hspace{2.0em}- **Negative and strict step-by-step ask-to-myself questions to prevent same mistake again**:

\hspace{3.0em}- Did I use the correct case for string literals?

\hspace{3.0em}- Did I ensure that the table names match the schema exactly?

\hspace{3.0em}- Did I verify that the query returns the required data accurately? \\

12. **Missing join for associated data**

\hspace{2.0em}- Question: What are the different cities where people live?

\hspace{2.0em}- **Incorrect SQL generated by me**: ```sql

\hspace{3.0em}SELECT DISTINCT city

\hspace{3.0em}FROM Addresses;

\hspace{3.0em}```

\hspace{2.0em}- **Corrected SQL generated by me**: ```sql

\hspace{3.0em}SELECT DISTINCT city FROM Addresses JOIN People\_Addresses ON Addresses.address\_id = People\_Addresses.address\_id;

\hspace{3.0em}```

\hspace{2.0em}- **Negative and strict step-by-step ask-to-myself questions to prevent same mistake again**:

\hspace{3.0em}- Did I join the necessary tables to ensure the data is associated correctly?

\hspace{3.0em}- Did I use the correct column names for the join condition?

\hspace{3.0em}- Did I verify that the query returns the required data accurately? \\

13. **Selecting unnecessary columns**

\hspace{2.0em}- Question: List all characteristics of product named ``sesame'' with type code ``Grade''.

\hspace{2.0em}- **Incorrect SQL generated by me**: ```sql

\hspace{3.0em}SELECT c.characteristic\_name, pc.product\_characteristic\_value

\hspace{3.0em}FROM Products p

\hspace{3.0em}JOIN Product\_Characteristics pc ON p.product\_id = pc.product\_id

\hspace{3.0em}JOIN Characteristics c ON pc.characteristic\_id = c.characteristic\_id

\hspace{3.0em}WHERE p.product\_name = 'sesame' AND c.characteristic\_type\_code = 'Grade';

\hspace{3.0em}```

\hspace{2.0em}- **Corrected SQL generated by me**: ```sql

\hspace{3.0em}SELECT c.characteristic\_name FROM Products p JOIN Product\_Characteristics pc ON p.product\_id = pc.product\_id JOIN Characteristics c ON pc.characteristic\_id = c.characteristic\_id WHERE p.product\_name = 'sesame' AND c.characteristic\_type\_code = 'Grade';

\hspace{3.0em}```

\hspace{2.0em}- **Negative and strict step-by-step ask-to-myself questions to prevent same mistake again**:

\hspace{3.0em}- Did I select only the required columns?

\hspace{3.0em}- Did I ensure that the query meets all specified requirements?

\hspace{3.0em}- Did I verify that the query returns the required data accurately? \\

14. **Case sensitivity in string literals**

\hspace{2.0em}- Question: Return the address and email of the customer with the first name Linda.

\hspace{2.0em}- **Incorrect SQL generated by me**: ```sql

\hspace{3.0em}SELECT a.address, c.email

\hspace{3.0em}FROM customer c

\hspace{3.0em}JOIN address a ON c.address\_id = a.address\_id

\hspace{3.0em}WHERE c.first\_name = 'LINDA';

\hspace{3.0em}```

\hspace{2.0em}- **Corrected SQL generated by me**: ```sql

\hspace{3.0em}SELECT a.address, c.email FROM customer c JOIN address a ON c.address\_id = a.address\_id WHERE c.first\_name = 'LINDA';

\hspace{3.0em}```

\hspace{2.0em}- **Negative and strict step-by-step ask-to-myself questions to prevent same mistake again**:

\hspace{3.0em}- Did I use the correct case for string literals?

\hspace{3.0em}- Did I ensure that the query returns the required data accurately? \\

15. **Incorrect string literals for city names**

\hspace{2.0em}- Question: What is the average price for flights from LA to Honolulu?

\hspace{2.0em}- **Incorrect SQL generated by me**: ```sql

\hspace{3.0em}SELECT AVG(price) AS average\_price

\hspace{3.0em}FROM flight

\hspace{3.0em}WHERE origin = 'LA' AND destination = 'Honolulu';

\hspace{3.0em}```

\hspace{2.0em}- **Corrected SQL generated by me**: ```sql

\hspace{3.0em}SELECT AVG(price) AS average\_price FROM flight WHERE origin = 'Los Angeles' AND destination = 'Honolulu';

\hspace{3.0em}```

\hspace{2.0em}- **Negative and strict step-by-step ask-to-myself questions to prevent same mistake again**:

\hspace{3.0em}- Did I use the correct city names as stored in the database?

\hspace{3.0em}- Did I use the correct syntax for string literals?

\hspace{3.0em}- Did I ensure that the query returns the required data accurately? \\

16. **Incorrect position value in query**

\hspace{2.0em}- Question: What are the names of all schools that have students trying out for the position of goal and 'mid'-field.

\hspace{2.0em}- **Incorrect SQL generated by me**: ```sql

\hspace{3.0em}SELECT cName

\hspace{3.0em}FROM Tryout

\hspace{3.0em}WHERE pPos = 'goal'

\hspace{3.0em}INTERSECT

\hspace{3.0em}SELECT cName

\hspace{3.0em}FROM Tryout

\hspace{3.0em}WHERE pPos = 'mid';

\hspace{3.0em}```

\hspace{2.0em}- **Successfully Corrected SQL using the feedback**: ```sql

\hspace{3.0em}SELECT cName FROM Tryout WHERE pPos = 'goalie' INTERSECT SELECT cName FROM Tryout WHERE pPos = 'mid';

\hspace{3.0em}```

\hspace{2.0em}- **Negative and strict step-by-step ask-to-myself questions to prevent same mistake again**:

\hspace{3.0em}- Did I use the correct position values as stored in the database?

\hspace{3.0em}- Did I ensure that the query returns the required data accurately? \\

17. **Unnecessary join and selecting unnecessary columns**

\hspace{2.0em}- Question: When do all the researcher role staff start to work, and when do they stop working?

\hspace{2.0em}- **Incorrect SQL generated by me**: ```sql

\hspace{3.0em}SELECT

\hspace{3.0em}ps.staff\_id,

\hspace{3.0em}ps.date\_from AS start\_date,

\hspace{3.0em}ps.date\_to AS end\_date

\hspace{3.0em}FROM

\hspace{3.0em}Project\_Staff ps

\hspace{3.0em}JOIN

\hspace{3.0em}Staff\_Roles sr ON ps.role\_code = sr.role\_code

\hspace{3.0em}WHERE

\hspace{3.0em}sr.role\_description = 'researcher';

\hspace{3.0em}```

\hspace{2.0em}- **Successfully Corrected SQL using the feedback**: ```sql

\hspace{3.0em}SELECT ps.date\_from AS start\_date, ps.date\_to AS end\_date FROM Project\_Staff ps WHERE ps.role\_code = 'researcher';

\hspace{3.0em}```

\hspace{2.0em}- **Negative and strict step-by-step ask-to-myself questions to prevent same mistake again**:

\hspace{3.0em}- Did I join unnecessary tables?

\hspace{3.0em}- Did I select only the required columns?

\hspace{3.0em}- Did I ensure that the query returns the required data accurately? \\

18. **Incorrect column name and unnecessary join**

\hspace{2.0em}- Question: What is the maximum training hours for the students whose training hours is greater than 1000 in different positions?

\hspace{2.0em}- **Incorrect SQL generated by me**: ```sql

\hspace{3.0em}SELECT pPos, MAX(HS) AS maxTrainingHours

\hspace{3.0em}FROM Player

\hspace{3.0em}JOIN Tryout ON Player.pID = Tryout.pID

\hspace{3.0em}WHERE HS > 1000

\hspace{3.0em}GROUP BY pPos;

\hspace{3.0em}```

\hspace{2.0em}- **Successfully Corrected SQL using the feedback**: ```sql

\hspace{3.0em}SELECT MAX(T1.HS) AS maxTrainingHours, T2.pPos FROM Player AS T1 JOIN Tryout AS T2 ON T1.pID = T2.pID WHERE T1.HS > 1000 GROUP BY T2.pPos;

\hspace{3.0em}```

\hspace{2.0em}- **Negative and strict step-by-step ask-to-myself questions to prevent same mistake again**:

\hspace{3.0em}- Did I use the correct column names?

\hspace{3.0em}- Did I use table aliases to improve readability?

\hspace{3.0em}- Did I ensure that the query returns the required data accurately? \\

19. **Incorrect column name and unnecessary join for company with most expensive furniture**

\hspace{2.0em}- Question: Find the id and number of shops for the company that produces the most expensive furniture.

\hspace{2.0em}- **Incorrect SQL generated by me**: ```sql

\hspace{3.0em}SELECT m.Manufacturer\_ID, m.Num\_of\_Shops

\hspace{3.0em}FROM manufacturer m

\hspace{3.0em}JOIN furniture\_manufacte fm ON m.Manufacturer\_ID = fm.Manufacturer\_ID

\hspace{3.0em}JOIN furniture f ON fm.Furniture\_ID = f.Furniture\_ID

\hspace{3.0em}WHERE f.Market\_Rate = (SELECT MAX(Market\_Rate) FROM furniture)

\hspace{3.0em}```

\hspace{2.0em}- **Successfully Corrected SQL using the feedback**: ```sql

\hspace{3.0em}SELECT m.Manufacturer\_ID, m.Num\_of\_Shops FROM manufacturer m JOIN furniture\_manufacte fm ON m.Manufacturer\_ID = fm.Manufacturer\_ID ORDER BY fm.Price\_in\_Dollar DESC LIMIT 1

\hspace{3.0em}```

\hspace{2.0em}- **Negative and strict step-by-step ask-to-myself questions to prevent same mistake again**:

\hspace{3.0em}- Did I join unnecessary tables?

\hspace{3.0em}- Did I use the correct column names for ordering?

\hspace{3.0em}- Did I ensure that the query is as efficient as possible?

\hspace{3.0em}- Did I verify that the query returns the required data accurately? \\

20. **Incorrect WHERE clause for least chargeable amount**

\hspace{2.0em}- Question: Which part has the least chargeable amount? List the part id and amount.

\hspace{2.0em}- **Incorrect SQL generated by me**: ```sql

\hspace{3.0em}SELECT part\_id, chargeable\_amount

\hspace{3.0em}FROM Parts

\hspace{3.0em}WHERE chargeable\_yn = 'Y'

\hspace{3.0em}ORDER BY chargeable\_amount ASC

\hspace{3.0em}LIMIT 1;

\hspace{3.0em}```

\hspace{2.0em}- **Successfully Corrected SQL using the feedback**: ```sql

\hspace{3.0em}SELECT part\_id, chargeable\_amount FROM Parts ORDER BY chargeable\_amount ASC LIMIT 1;

\hspace{3.0em}```

\hspace{2.0em}- **Negative and strict step-by-step ask-to-myself questions to prevent same mistake again**:

\hspace{3.0em}- Did I use the correct conditions in the WHERE clause?

\hspace{3.0em}- Did I ensure that the query considers all parts as required?

\hspace{3.0em}- Did I verify that the query returns the required data accurately? \\

21. **Grouping by department name instead of department code**

\hspace{2.0em}- Question: Find the name of the department that offers the largest number of credits of all classes.

\hspace{2.0em}- **Incorrect SQL generated by me**: ```sql

\hspace{3.0em}SELECT d.DEPT\_NAME, SUM(c.CRS\_CREDIT) AS total\_credits

\hspace{3.0em}FROM CLASS cl

\hspace{3.0em}JOIN COURSE c ON cl.CRS\_CODE = c.CRS\_CODE

\hspace{3.0em}JOIN DEPARTMENT d ON c.DEPT\_CODE = d.DEPT\_CODE

\hspace{3.0em}GROUP BY d.DEPT\_NAME

\hspace{3.0em}ORDER BY total\_credits DESC

\hspace{3.0em}LIMIT 1;

\hspace{3.0em}```

\hspace{2.0em}- **Successfully Corrected SQL using the feedback**: ```sql

\hspace{3.0em}SELECT d.dept\_name FROM CLASS AS cl JOIN COURSE AS c ON cl.crs\_code = c.crs\_code JOIN DEPARTMENT AS d ON c.dept\_code = d.dept\_code GROUP BY d.dept\_code ORDER BY SUM(c.crs\_credit) DESC LIMIT 1;

\hspace{3.0em}```

\hspace{2.0em}- **Negative and strict step-by-step ask-to-myself questions to prevent same mistake again**:

\hspace{3.0em}- Did I join the correct tables to verify the functional\_area\_description?

\end{tiny}
\end{mybox}

\begin{mybox}[label={qwen32b-bird}]{Guidelines for Qwen2.5-Coder-32B-Instruct on BIRD}
\begin{tiny}

1. **Not selecting the description field for body height status**

\hspace{1.5em}- Question: ``Give the body height status of Mr. Vincent Wyman on 2010/8/2.''

\hspace{1.5em}- **Incorrect SQL generated by me**: ```Incorrect corrected sql ```

\hspace{1.5em}- **Corrected SQL generated by me**: ```sql SELECT T2.DESCRIPTION, T2.VALUE, T2.UNITS FROM patients AS T1 INNER JOIN observations AS T2 ON T1.patient = T2.PATIENT WHERE T1.prefix = 'Mr.' AND T1.first = 'Vincent' AND T1.last = 'Wyman' AND T2.DATE = '2010-08-02' AND T2.DESCRIPTION = 'Body Height'```

\hspace{1.5em}- **Negative and strict step-by-step ask-to-myself questions to prevent same mistake again**:

\hspace{2.5em}- Did I include all necessary fields in the SELECT clause?

\hspace{2.5em}- Have I verified that the DESCRIPTION field is required for the output?

\hspace{2.5em}- Am I correctly joining the tables to get the required data? \\

2. **Incorrect join condition for identifying the captain of the winning team**

\hspace{1.5em}- Question: ``Who was the captain of the winning team in the match held on 1st June 2008?''

\hspace{1.5em}- **Incorrect SQL generated by me**: ```SELECT T3.Player\_Name FROM `Match` AS T1 INNER JOIN Player\_Match AS T2 ON T1.Match\_Winner = T2.Team\_Id INNER JOIN Player AS T3 ON T2.Player\_Id = T3.Player\_Id INNER JOIN Rolee AS T4 ON T2.Role\_Id = T4.Role\_Id WHERE T1.Match\_Date = '2008-06-01' AND T4.Role\_Desc = 'Captain'```

\hspace{1.5em}- **Corrected SQL generated by me**: ```sql SELECT T3.Player\_Name FROM `Match` AS T1 INNER JOIN Player\_Match AS T2 ON T1.Match\_Id = T2.Match\_Id INNER JOIN Player AS T3 ON T2.Player\_Id = T3.Player\_Id INNER JOIN Rolee AS T4 ON T2.Role\_Id = T4.Role\_Id WHERE T1.Match\_Date = '2008-06-01' AND T2.Team\_Id = T1.Match\_Winner AND T4.Role\_Desc = 'Captain'```

\hspace{1.5em}- **Negative and strict step-by-step ask-to-myself questions to prevent same mistake again**:

\hspace{2.5em}- Is the join condition based on the correct unique identifier?

\hspace{2.5em}- Have I ensured that the join order is logical and correct?

\hspace{2.5em}- Am I correctly filtering the data to match the question's requirements? \\

3. **Not casting count to a real number for percentage calculation**

\hspace{1.5em}- Question: ``What percentage of students enlisted in the navy are male?''

\hspace{1.5em}- **Incorrect SQL generated by me**: ```SELECT CAST(COUNT(CASE WHEN T1.organ = 'navy' THEN T2.name ELSE NULL END) AS REAL) * 100 / COUNT(T2.name) FROM enlist AS T1 INNER JOIN male AS T2 ON T1.`name` = T2.`name` WHERE T1.organ = 'navy'```

\hspace{1.5em}- **Corrected SQL generated by me**: ```sql SELECT CAST(COUNT(T2.name) AS REAL) * 100 / COUNT(T1.name) FROM enlist AS T1 LEFT JOIN male AS T2 ON T1.`name` = T2.`name` WHERE T1.organ = 'navy'```

\hspace{1.5em}- **Negative and strict step-by-step ask-to-myself questions to prevent same mistake again**:

\hspace{2.5em}- Have I cast the count to a real number to ensure a decimal result?

\hspace{2.5em}- Am I using the correct type of join for the data retrieval?

\hspace{2.5em}- Have I verified that the logic in the CASE statement is correct?

\hspace{2.5em}- Am I dividing by the correct count to get the percentage? \\

4. **Using SUM and CASE statement unnecessarily for listing full names**

\hspace{1.5em}- Question: ``List the full name of all the 'Production Technician - WC50'''

\hspace{1.5em}- **Incorrect SQL generated by me**: ```SELECT SUM(T1.Likes) FROM twitter AS T1 INNER JOIN location AS T2 ON T2.LocationID = T1.LocationID WHERE T1.Sentiment > 0```

\hspace{1.5em}- **Corrected SQL generated by me**: ```sql SELECT T2.FirstName, T2.MiddleName, T2.LastName FROM Employee AS T1 INNER JOIN Person AS T2 ON T1.BusinessEntityID = T2.BusinessEntityID WHERE T1.JobTitle = 'Production Technician - WC50'```

\hspace{1.5em}- **Negative and strict step-by-step ask-to-myself questions to prevent same mistake again**:

\hspace{2.5em}- Have I selected the correct tables and columns related to the task?

\hspace{2.5em}- Am I using the correct join condition?

\hspace{2.5em}- Have I verified that the filtering condition matches the question's requirements?

\hspace{2.5em}- Am I using COUNT or SUM unnecessarily for listing data? \\

5. **Calculating percentage instead of average for object samples**

\hspace{1.5em}- Question: ``Calculate the average of object samples for the image.''

\hspace{1.5em}- **Incorrect SQL generated by me**: ```SELECT (SUM(CASE WHEN gender = 'M' THEN 1 ELSE 0 END) * 100.0 / COUNT(device\_id)) AS male\_percentage FROM gender\_age;```

\hspace{1.5em}- **Corrected SQL generated by me**: ```sql SELECT AVG(obj\_count) AS average\_object\_samples\_per\_image FROM ( SELECT IMG\_ID, COUNT(OBJ\_SAMPLE\_ID) AS obj\_count FROM IMG\_OBJ GROUP BY IMG\_ID ) AS image\_obj\_counts;```

\hspace{1.5em}- **Negative and strict step-by-step ask-to-myself questions to prevent same mistake again**:

\hspace{2.5em}- Have I used the correct table and columns related to the task?

\hspace{2.5em}- Am I calculating the correct metric (average instead of percentage)?

\hspace{2.5em}- Have I verified that the division is done by the correct count?

\hspace{2.5em}- Am I using the correct syntax for subqueries? \\

6. **Using LIKE instead of equals for exact match and not using COUNT**

\hspace{1.5em}- Question: ``How many 'cool' type compliments does user No. 41717 get?''

\hspace{1.5em}- **Incorrect SQL generated by me**: ```SELECT T2.number\_of\_compliments FROM Compliments AS T1 INNER JOIN Users\_Compliments AS T2 ON T1.compliment\_id = T2.compliment\_id WHERE T2.user\_id = 41717 AND T1.compliment\_type LIKE 'cool'```

\hspace{1.5em}- **Corrected SQL generated by me**: ```sql SELECT COUNT(*) FROM Compliments AS T1 INNER JOIN Users\_Compliments AS T2 ON T1.compliment\_id = T2.compliment\_id WHERE T2.user\_id = 41717 AND T1.compliment\_type = 'cool'```

\hspace{1.5em}- **Negative and strict step-by-step ask-to-myself questions to prevent same mistake again**:

\hspace{2.5em}- Am I using the correct operator for exact matches?

\hspace{2.5em}- Have I used COUNT to get the number of rows that match the criteria?

\hspace{2.5em}- Have I verified that the join condition is correct?

\hspace{2.5em}- Am I selecting the correct fields for the output? \\

7. **Selecting the wrong column and filtering by the wrong condition**

\hspace{1.5em}- Question: ``What is the image ID with a predicted class of 'parked on'?''

\hspace{1.5em}- **Incorrect SQL generated by me**: ```SELECT timestamp FROM events WHERE event\_id = 12;```

\hspace{1.5em}- **Corrected SQL generated by me**: ```sql SELECT T2.IMG\_ID FROM IMG\_REL AS T1 INNER JOIN PRED\_CLASSES AS T2 ON T1.PRED\_CLASS\_ID = T2.PRED\_CLASS\_ID WHERE T2.PRED\_CLASS = 'parked on'```

\hspace{1.5em}- **Negative and strict step-by-step ask-to-myself questions to prevent same mistake again**:

\hspace{2.5em}- Have I selected the correct column for the output?

\hspace{2.5em}- Am I filtering the data by the correct condition?

\hspace{2.5em}- Have I used the correct tables for the query?

\hspace{2.5em}- Am I using the correct join condition? \\

8. **Grouping by non-unique identifier**

\hspace{1.5em}- Question: ``State the name of the city with the most venues.''

\hspace{1.5em}- **Incorrect SQL generated by me**: ```SELECT T1.City\_Name FROM City AS T1 INNER JOIN Venue AS T2 ON T1.City\_Id = T2.City\_Id GROUP BY T1.City\_Name ORDER BY COUNT(T2.Venue\_Id) DESC LIMIT 1```

\hspace{1.5em}- **Corrected SQL generated by me**: ```sql SELECT T1.City\_Name FROM City AS T1 INNER JOIN Venue AS T2 ON T1.City\_Id = T2.City\_Id GROUP BY T1.City\_Id ORDER BY COUNT(T2.Venue\_Id) DESC LIMIT 1```

\hspace{1.5em}- **Negative and strict step-by-step ask-to-myself questions to prevent same mistake again**:

\hspace{2.5em}- Am I grouping by the correct unique identifier?

\hspace{2.5em}- Have I verified that the join condition is correct?

\hspace{2.5em}- Have I ensured that the filtering conditions match the question's requirements?

\hspace{2.5em}- Am I selecting the correct fields for the output? \\

9. **Selecting wrong tables and columns, using incorrect join condition, and filtering with irrelevant criteria**

\hspace{1.5em}- Question: ``List down the titles and descriptions of the crimes cases against persons.''

\hspace{1.5em}- **Incorrect SQL generated by me**: ```SELECT T2.AggregationMethod FROM Indicators AS T1 INNER JOIN Series AS T2 ON T1.IndicatorName = T2.SeriesCode WHERE T1.CountryName = 'Arab World' AND T1.Year = 1960 AND T1.Value = 133```

\hspace{1.5em}- **Corrected SQL generated by me**: ```sql SELECT title, description FROM FBI\_Code WHERE crime\_against = 'Persons'```

\hspace{1.5em}- **Negative and strict step-by-step ask-to-myself questions to prevent same mistake again**:

\hspace{2.5em}- Have I selected the correct tables and columns related to the task?

\hspace{2.5em}- Am I using the correct join condition?

\hspace{2.5em}- Have I verified that the filtering conditions match the question's requirements?

\hspace{2.5em}- Am I selecting the correct fields for the output? \\

10. **Selecting names instead of counting players who became coaches and are in the Hall of Fame**

\hspace{2.0em}- Question: ``Among the players who became coaches, how many of them have gotten in the Hall of Fame?''

\hspace{2.0em}- **Incorrect SQL generated by me**: ```SELECT T2.name FROM Award AS T1 INNER JOIN Person AS T2 ON T1.person\_id = T2.person\_id WHERE T1.award\_id = 313```

\hspace{2.0em}- **Corrected SQL generated by me**: ```sql SELECT COUNT(*) FROM Master AS T1 INNER JOIN HOF AS T2 ON T1.hofID = T2.hofID WHERE T1.playerID IS NOT NULL AND T1.coachID IS NOT NULL```

\hspace{2.0em}- **Negative and strict step-by-step ask-to-myself questions to prevent same mistake again**:

\hspace{3.0em}- Have I selected the correct tables and columns related to the task?

\hspace{3.0em}- Am I using the correct join condition?

\hspace{3.0em}- Have I verified that the filtering conditions match the question's requirements?

\hspace{3.0em}- Am I selecting the correct fields for the output?

\hspace{3.0em}- Am I using COUNT to get the number of players who meet the criteria? \\

11. **Selecting wrong tables and columns for group discounts for resellers**

\hspace{2.0em}- Question: ``What categories of offers qualify for group discounts for resellers?''

\hspace{2.0em}- **Incorrect SQL generated by me**: ```SELECT T1.email FROM student AS T1 INNER JOIN registration AS T2 ON T1.student\_id = T2.student\_id INNER JOIN course AS T3 ON T2.course\_id = T3.course\_id WHERE T2.grade = 'B' AND T3.diff > ( SELECT AVG(diff) * 0.8 FROM course )```

\hspace{2.0em}- **Corrected SQL generated by me**: ```sql SELECT Type FROM SpecialOffer WHERE Category = 'Reseller'```

\hspace{2.0em}- **Negative and strict step-by-step ask-to-myself questions to prevent same mistake again**:

\hspace{3.0em}- Did I use the correct table for the query?

\hspace{3.0em}- Did I correctly specify the columns to select?

\hspace{3.0em}- Did I use the correct filtering condition?

\hspace{3.0em}- Did I unnecessarily join multiple tables?

\hspace{3.0em}- Have I ensured that my conditions accurately target the required data without adding unnecessary complexity?

\hspace{3.0em}- Did I use the correct syntax for string literals? \\

12. **Incorrectly using a subquery to find the sales representative**

\hspace{2.0em}- Question: ``How many customers have an employee who reports to William Patterson as their sales representative?''

\hspace{2.0em}- **Incorrect SQL generated by me**: ```SELECT COUNT(T1.customerNumber) FROM customers AS T1 INNER JOIN employees AS T2 ON T1.salesRepEmployeeNumber = T2.employeeNumber WHERE T2.reportsTo = ( SELECT employeeNumber FROM employees WHERE firstName = 'William' AND lastName = 'Patterson' )```

\hspace{2.0em}- **Corrected SQL generated by me**: ```sql SELECT COUNT(T1.customerNumber) FROM customers AS T1 INNER JOIN employees AS T2 ON T1.salesRepEmployeeNumber = T2.employeeNumber WHERE T2.firstName = 'William' AND T2.lastName = 'Patterson'```

\hspace{2.0em}- **Negative and strict step-by-step ask-to-myself questions to prevent same mistake again**:

\hspace{3.0em}- Have I selected the correct tables and columns related to the task?

\hspace{3.0em}- Am I using the correct join condition?

\hspace{3.0em}- Have I verified that the filtering conditions match the question's requirements?

\hspace{3.0em}- Am I selecting the correct fields for the output?

\hspace{3.0em}- Am I using subqueries unnecessarily? \\

13. **Selecting the wrong table and columns for currency code**

\hspace{2.0em}- Question: ``Give the Mauritius Rupee's currency code.''

\hspace{2.0em}- **Incorrect SQL generated by me**: ```SELECT T2.Name FROM characters AS T1 INNER JOIN actor AS T2 ON T1.ActorID = T2.ActorID WHERE T1.`Character Name` = 'Chanice Kobolowski';```

\hspace{2.0em}- **Corrected SQL generated by me**: ```sql SELECT CurrencyCode FROM Currency WHERE Name = 'Mauritius Rupee'```

\hspace{2.0em}- **Negative and strict step-by-step ask-to-myself questions to prevent same mistake again**:

\hspace{3.0em}- Have I selected the correct tables and columns related to the task?

\hspace{3.0em}- Am I using the correct join condition?

\hspace{3.0em}- Have I verified that the filtering conditions match the question's requirements?

\hspace{3.0em}- Am I selecting the correct fields for the output? \\

14. **Incorrectly calculating the percentage of trips done by subscribers**

\hspace{2.0em}- Question: ``What is the percentage of the trips were done by a subscriber?''

\hspace{2.0em}- **Incorrect SQL generated by me**: ```SELECT CustomerID, `Customer Names` FROM Customers WHERE `Customer Names` LIKE 'W\%'```

\hspace{2.0em}- **Corrected SQL generated by me**: ```sql SELECT CAST(COUNT(CASE WHEN subscription\_type = 'Subscriber' THEN id ELSE NULL END) AS REAL) * 100 / COUNT(*) FROM trip```

\hspace{2.0em}- **Negative and strict step-by-step ask-to-myself questions to prevent same mistake again**:

\hspace{3.0em}- Have I selected the correct tables and columns related to the task?

\hspace{3.0em}- Am I using the correct join condition?

\hspace{3.0em}- Have I verified that the filtering conditions match the question's requirements?

\hspace{3.0em}- Am I selecting the correct fields for the output?

\hspace{3.0em}- Am I using the correct syntax for division to get a decimal result?

\hspace{3.0em}- Am I counting the correct rows for the percentage calculation? \\

15. **Incorrectly selecting OrderNumber instead of order names**

\hspace{2.0em}- Question: ``List out the name of orders which have delivery date of 6/13/2018.''

\hspace{2.0em}- **Incorrect SQL generated by me**: ```SELECT OrderNumber FROM `Sales Orders` WHERE DeliveryDate = '6/13/2018';```

\hspace{2.0em}- **Corrected SQL generated by me**: ```sql SELECT OrderNumber FROM `Sales Orders` WHERE DeliveryDate = '2018-06-13';```

\hspace{2.0em}- **Negative and strict step-by-step ask-to-myself questions to prevent same mistake again**:

\hspace{3.0em}- Have I selected the correct tables and columns related to the task?

\hspace{3.0em}- Am I using the correct join condition?

\hspace{3.0em}- Have I verified that the filtering conditions match the question's requirements?

\hspace{3.0em}- Am I selecting the correct fields for the output?

\hspace{3.0em}- Am I using the correct date format for the filtering condition? \\

16. **Case sensitivity issue in county names**

\hspace{2.0em}- Question: ``List all the cities in Sonoma County.''

\hspace{2.0em}- **Incorrect SQL generated by me**: ```SELECT city FROM geographic WHERE county = 'Sonoma County';```

\hspace{2.0em}- **Corrected SQL generated by me**: ```sql SELECT city FROM geographic WHERE county = 'sonoma county';```

\hspace{2.0em}- **Negative and strict step-by-step ask-to-myself questions to prevent same mistake again**:

\hspace{3.0em}- Have I selected the correct tables and columns related to the task?

\hspace{3.0em}- Am I using the correct join condition?

\hspace{3.0em}- Have I verified that the filtering conditions match the question's requirements?

\hspace{3.0em}- Am I selecting the correct fields for the output?

\hspace{3.0em}- Am I using the correct case for string literals to match the database schema? \\

17. **Selecting from wrong tables and using incorrect filtering conditions**

\hspace{2.0em}- Question: ``Give the level of education and occupation of customers ages from 20 to 35 with an income K of 2000 and below.''

\hspace{2.0em}- **Incorrect SQL generated by me**: ```SELECT T1.Seriescode FROM FootNotes AS T1 INNER JOIN Series AS T2 ON T1.Seriescode = T2.SeriesCode WHERE T1.Year LIKE \'\%2005\%\' AND T2.Source = 'International Monetary Fund, Balance of Payments Statistics Yearbook and data files.'```

\hspace{2.0em}- **Corrected SQL generated by me**: ```sql SELECT T1.EDUCATIONNUM, T1.OCCUPATION FROM Customers AS T1 INNER JOIN Demog AS T2 ON T1.GEOID = T2.GEOID WHERE T1.age BETWEEN 20 AND 35 AND T2.INCOME\_K <= 2000```

\hspace{2.0em}- **Negative and strict step-by-step ask-to-myself questions to prevent same mistake again**:

\hspace{3.0em}- Have I selected the correct tables and columns related to the task?

\hspace{3.0em}- Am I using the correct join condition?

\hspace{3.0em}- Have I verified that the filtering conditions match the question's requirements?

\hspace{3.0em}- Am I selecting the correct fields for the output? \\

18. **Incorrectly selecting match IDs and filtering by unrelated conditions**

\hspace{2.0em}- Question: ``Please list all horror films that have a rating of 1.''

\hspace{2.0em}- **Incorrect SQL generated by me**: ```SELECT DISTINCT m.movieid FROM movies2directors m2d JOIN u2base u```

\hspace{2.0em}- **Corrected SQL generated by me**: ```sql SELECT m.movieid FROM movies m JOIN u2base u ON m.movieid = u.movieid JOIN movies2directors md ON m.movieid = md.movieid WHERE md.genre = 'Horror' AND u.rating = 1;```

\hspace{2.0em}- **Negative and strict step-by-step ask-to-myself questions to prevent same mistake again**:

\hspace{3.0em}- Have I selected the correct tables and columns related to the task?

\hspace{3.0em}- Am I using the correct join condition?

\hspace{3.0em}- Have I verified that the filtering conditions match the question's requirements?

\hspace{3.0em}- Am I selecting the correct fields for the output?

\hspace{3.0em}- Am I using `DISTINCT` unnecessarily?

\end{tiny}
\end{mybox}
\end{appendix}

\end{document}